\documentclass[letterpaper, 10 pt, conference]{ieeeconf}
\IEEEoverridecommandlockouts%

\usepackage{amsmath}
\usepackage{amssymb}
\usepackage{interval}
\usepackage{graphicx}
\usepackage[export]{adjustbox}
\usepackage{subcaption}
\usepackage{mdframed}
\usepackage{array}
\usepackage{cite}
\usepackage{balance}
\usepackage[utf8]{inputenc}
\usepackage[textsize=scriptsize]{todonotes}

\intervalconfig{soft open fences}
\usepackage[noabbrev,capitalise]{cleveref}
\usepackage[gen]{eurosym}

\newcommand\copyrighttext{%
  \footnotesize \textcopyright 2019 IEEE.  Personal use of this material is permitted.  Permission from IEEE must be obtained for all other uses, in any current or future media, including reprinting/republishing this material for advertising or promotional purposes, creating new collective works, for resale or redistribution to servers or lists, or reuse of any copyrighted component of this work in other works.}
\newcommand\copyrightnotice{%
\begin{tikzpicture}[remember picture,overlay]
\node[anchor=south,yshift=10pt] at (current page.south) {\fbox{\parbox{\dimexpr\textwidth-\fboxsep-\fboxrule\relax}{\copyrighttext}}};
\end{tikzpicture}%
}

\title{\LARGE \textbf{%
Robots that Sync and Swarm: A Proof of Concept in ROS\,2}}

\author{Agata Barci\'{s}, Micha\l{} Barci\'{s} and Christian Bettstetter
\thanks{All authors are associated with the University of Klagenfurt, Austria. Agata~Barci\'s
and Micha\l~Barci\'s are associated with the Karl Popper School on Networked
Autonomous Aerial Vehicles. Christian Bettstetter is associated with the Institute of
Networked and Embedded Systems.}}

\begin{document}

\maketitle
\copyrightnotice

\begin{abstract}
    A unified mathematical model for synchronisation and swarming has recently been
    proposed. Each system entity, called a ``swarmalator'', coordinates
    its internal phase and location with the other entities in a way that these
    two attributes are mutually coupled. This paper realises and studies, for
    the first time, the concept of swarmalators in a technical system. We adapt
    and extend the original model for its use with mobile robots and implement it
    in the Robot Operating System 2 (ROS\,2). Simulations and experiments with
    small robots demonstrate the feasibility of the model and show its potential
    to be applied to real-world systems. All types of space-time patterns
    achieved in theory can be reproduced in practice. Applications can be found
    in monitoring, exploration, entertainment and art, among other~domains.
\end{abstract}


\newcommand{\x}{{\mathbf{x}}}
\newcommand{\vd}{{\mathbf{v}}^{(\textrm d)}}
\newcommand{\dist}{d}
\newcommand{\phase}{\phi}
\newcommand{\orient}{\theta}
\newcommand{\numnodes}{N}
\newcommand{\J}{J}
\newcommand{\direction}{\mathbf{\hat{n}}}
\newcommand{\attpos}{\mathbf{I}_1}
\newcommand{\reppos}{\mathbf{I}_2}
\newcommand{\phasepos}{F}
\newcommand{\attphase}{H}
\newcommand{\posphase}{G_{\phase}}
\newcommand{\attorient}{R}
\newcommand{\posorient}{G_{\orient}}
\newcommand{\velorient}{S}
\newcommand{\dimnumber}{m}

\newcommand{\norm}[1]{\left\lVert#1\right\rVert}

\definecolor{myblue}{rgb}{0,0.35,0.9}
\mdfdefinestyle{theoremstyle}{%
linecolor=black,linewidth=1pt,%
innertopmargin=\topskip,
}
\mdtheorem[style=theoremstyle]{framed-box}{Box}

\IEEEpeerreviewmaketitle%

\section{Introduction}

Two exciting phenomena of self-organisation occurring in natural and technical
systems are synchronisation and swarming (see~\cite{Bonabeau:1999:NAS:554879,camazine_self-organization_2003,strogatz_sync_2004,hamann_swarm_2018}).
In simple words, synchronisation is coordination in time and swarming is
coordination in space. Scientific work on these phenomena remained mostly
disconnected until O'Keeffe, Hong and Strogatz 
proposed a unified mathematical model for entities that both
synchronise and swarm. These entities, called \textit{swarmalators}, are ``oscillators whose phase
dynamics and spatial dynamics are coupled''~\cite{okeeffe_oscillators_2017}.
A swarmalator's phase dynamics results from location-dependent synchronisation and its
spatial dynamics results from phase-dependent aggregation. Groups of
swarmalators show visually appealing spatio-temporal patterns, including phase waves and cluster~formations.

The objective of our work is to transfer and adapt the
swarmalator model to mobile robotics and employ it in real-world systems. In fact, we see many applications for swarmalators in robotics. In
monitoring and surveillance, the model can be used to coordinate robots during a
patrolling mission or gather them around a point
of interest. Depending on the specific tasks, the coupling of phase and position can be
utilised in different ways. For instance, robots can take consecutive pictures of
the point of interest from different viewing angles, or position-dependent
communication slots can be assigned. In exploration, a group of underwater
vehicles can swim in a formation with the movements of their fins synchronised
in order to improve performance. In art and entertainment, swarmalators can draw
artificial paintings and perform aerial light shows with~drones. 

As a first step towards robotic swarmalators, we present the following contributions: we
implement the swarmalator model in the Robot Operating System 2 (ROS\,2),
reproduce spatio-temporal patterns of~\cite{okeeffe_oscillators_2017}, modify
and extend the model to account for the specific movement properties of robots
(namely collision avoidance and constraints in the movement directions), propose
a new swarming platform based on Balboa robots~\cite{noauthor_pololu_nodate} and
finally use this platform to present a proof of concept demonstrating for the
first time the feasibility of real robots acting as swarmalators. ROS\,2 is
used instead of ROS because it operates without a central unit, which
makes it suited for swarm robotics.

The paper is structured as follows: \cref{section:system-model} presents a
swarmalator model for robots.  \Cref{sec:impl} briefly describes the robot
platform and the implementation in ROS\,2. \Cref{sec:results} shows the results
from simulations and experiments. \Cref{sec:rel-work} covers related work.
Finally, \Cref{sec:concl} concludes.

\section{Swarmalator Model}\label{section:system-model}

\subsection{The Original Model}

A swarmalator $i$ has a position $\x_{i} \in \mathbb{R}^\dimnumber$ in the
$\dimnumber$-dimensional space and an internal phase $\phase_{i} \in
\interval[open right]{0}{2\pi}$.  Its natural frequency of oscillation is
$\omega_i$. The position difference between two entities $i$ and $j$ is
$\x_{ij}=\x_{j}-\x_{i}$ and their phase difference is
$\phase_{ij}=\phase_{j}-\phase_{i}$. The behaviour of entity~$i$ in a system of
$\numnodes$ entities is modelled by two differential
equations~\cite{okeeffe_oscillators_2017}: Phase-dependent movement is given by 
\begin{equation}\label{eq:model-x} 
    \dot{\x}_i = \frac{1}{\numnodes} \sum_{j\neq i}^{\numnodes} \big[%
    \attpos(\x_{ij}) \: \phasepos(\phase_{ij}) - \reppos(\x_{ij}) \big], 
\end{equation} 
and position-dependent synchronisation is given by
\begin{equation}\label{eq:model-phase} 
    \dot{\phase}_i = \omega_i + \frac{K}{\numnodes} \sum_{j\neq i}^{\numnodes} 
    \attphase(\phase_{ij}) \: \posphase(\x_{ij}).
\end{equation}
The function $\attpos$ describes the spatial attraction between two entities and
$\phasepos$ determines how this attraction is influenced by their phase
similarity. $\reppos$ describes the spatial repulsion between two entities. In a
similar way, the synchronisation is defined by two functions: $\attphase$ is the
attraction of the phases and $\posphase$ is its dependency on the proximity of
two entities.

For swarmalators in two dimensions ($\dimnumber\!=\!2$), the following functions
are used in~\cite{okeeffe_oscillators_2017}: $\attpos(\x_{ij}) = \frac{\x_{ij}}{\norm{\x_{ij}}}$, $
    \reppos(\x_{ij}) = \frac{\x_{ij}}{\norm{\x_{ij}}^2}$, $\phasepos(\phase_{ij}) = 1 + \J \cos \phase_{ij}$, $\attphase(\phase_{ij}) = \sin \phase_{ij}$, $\posphase(\x_{ij}) = \frac{1}{\norm{\x_{ij}}}$.

This system has two parameters: $J$ and $K$. The parameter $J \in \interval
{-1}{1}$ determines how strong the influence of phase similarity on spatial
attraction is. If $J$ is positive, an entity wants to be close to entities with
the same phase. If $J$ is negative, an entity is attracted by entities with the
opposite phase. The original model does not account for the orientation of
entities. The parameter $K\!\in\! \mathbb{R}$ determines how strongly coupled
the phases of two entities are.  Positive values of this coupling strength tend
to lower the phase difference between entities. Negative values tend to
increase~it. 

Applying this model to a set of entities yields, after some time, different
patterns depending on the choice of parameters.
These patterns can be classified into five categories: static sync, static
async, static phase wave, splintered phase wave and active phase
wave~\cite{okeeffe_oscillators_2017}. Such swarmalator patterns are analysed in more detail in \cite{okeeffe_ring_2018} and \cite{hong18:chaos}. All this work is however purely analytical and does no take into account the specific characteristics of robots. In order to apply the swarmalator model in robotics, we have to perform some modifications and extensions, as described in the following.

\subsection{A Model Suited for Robots}

\subsubsection{Modifications Due to Physical Constraints}

The Balboa robots have the following physical limitations:

\begin{itemize}
    \item \textit{Movement constraints:} A swarmalator in the original model can
        move freely in all directions. In contrast, the robot can only turn
        around its centre or move in the direction it is facing, either forward
        or backward. It is therefore impossible to execute the velocity
        behaviour as described in \Cref{eq:model-x}. Instead, we interpret it as
        the \textit{desired velocity}, denoted by $\vd_i$ for robot $i$.
        Additionally, we introduce a new state variable~$\orient_i$ for the
        orientation of the robot. The function~$\velorient$ describes how the
        robot's orientation is influenced by its desired velocity. Finally, the
        velocity $\dot{\x}_i$ is the component of the desired velocity in the
        direction that the robot is facing. All equations are given in
        Box~\ref{box}.

    \item \textit{Speed limit:} The speed of the robot is limited
        by its drivetrain. The linear and angular velocities obtained from the
        equations need to be limited appropriately.

    \item \textit{Collision avoidance:} A swarmalator is
        treated as a point-shaped entity whereas robots occupy a certain
        area. To prevent robots from colliding with each other, we define a
        circular safety area around each robot. The function $\reppos$ is
        redefined using the distance between these safety areas, denoted by
        $\dist_{ij}$. This ensures that the robots will repel each other if they
        get too close.
\end{itemize}

\subsubsection{Alignment of Orientations}

We also introduce an extension to the swarmalator model in which the
orientations $\orient$ can be aligned. The function $\attorient$ describes how
the orientations of two entities are attracted. The function $\posorient$
determines how their spatial proximity influences this attraction. The
attraction is controlled by the~coefficient 
\begin{equation}
    \lambda = \min \left\{1, \frac{\|\vd_i\|}{PC} \right\} 
\end{equation}
with the parameter $P \in \mathbb{R}^+_0$ defining the attraction strength 
and the constant $C \in \mathbb{R}^+$  depending on the movement specifics.  If
$P = 0$, the entity will always turn to the direction of its desired movement.
As $P$ increases, the influence from other entities increases and finally, for
$P \to \infty$, the entity will align only with its neighbours and never
try to turn in the direction of the desired velocity.

The product $PC$ can be seen as a threshold velocity above which an entity does
not align with other entities but only with its desired velocity. A particular
value is $P = 1$ because the threshold velocity becomes equal to $C$.  In this
paper, $C$ is set to be the maximum velocity of the robots. If the model tries
to make an entity move faster than possible, this entity will only turn in the
desired direction and remain unaffected by the orientations of other entities.
As soon as the desired velocity drops and is achievable, the orientation will
start to be influenced by others. At the full stop, only the orientation
attraction will be considered.
\smallskip 
\begin{framed-box}[Swarmalator model for mobile robots]\label{box}
\begin{equation*}\label{eq:ext-model-vd} 
    \vd_i = \frac{1}{\numnodes} \sum_{j\neq i}^{\numnodes} \left[%
    \attpos(\x_{ij}) \phasepos(\phase_{ij}) - \reppos(\x_{ij}) \right]
\end{equation*}
\begin{equation*}\label{eq:ext-model-orient} 
    \dot{\orient}_i = (1 - \lambda) \frac{1}{\numnodes}
    \sum_{j\neq i}^{\numnodes} \attorient(\orient_{ij}) \posorient(\x_{ij})
    + \lambda \velorient(\orient_{i}, \vd_i)
\end{equation*}
\begin{equation*}\label{eq:ext-model-x}
    \dot{\x}_i = \vd_i \cos{\orient_i}
\end{equation*} 
\begin{equation*}\label{eq:ext-model-phase} 
    \dot{\phase}_i = \omega_i + \frac{K}{\numnodes} \sum_{j\neq i}^{\numnodes} 
    \attphase(\phase_{ij}) \posphase(\x_{ij})
\end{equation*}
\end{framed-box}
\smallskip 

The overall model is given in Box~\ref{box}. Entities that synchronise and swarm
according to this model are called \textit{swarmalatorbots}.
We use $\attpos$, $\phasepos$, $\attphase$ and $\posphase$ from the original
model described above. Orientation attraction is chosen to be the same as
phase attraction: 
\begin{equation}\label{eq:ext-func-ratt}
    \attorient(\orient_{ij}) = \sin \orient_{ij},
\end{equation} 
\begin{equation}\label{eq:ext-func-gorient}
\posorient(\x_{ij}) = \frac{1}{\norm{\x_{ij}}}.\end{equation}
Furthermore, we use: 
\begin{equation}\label{eq:ext-func-irep}
    \reppos(\x_{ij}) = \frac{\x_{ij}}{\dist_{ij}^2},
\end{equation}
\begin{equation}\label{eq:ext-func-s}
    \velorient(\orient_i, \vd_i) = \sin(\angle(\vd_i) - \orient_i).
\end{equation}

\section{Implementation}\label{sec:impl}

This section describes the robot platform used, the implementation of the
swarmalatorbots in ROS\,2 and other modules required to perform experiments. 

\subsection{Mobile Robot Platform}\label{sec:impl-hardware}

Multiple platforms are available for swarm robotics, but many of them have very limited computational power (e.g.~Kilobots~\cite{rubenstein_kilobot_2012} and
Spiderinos~\cite{jdeed_spiderino_2017}). For this work, we decided to prepare a platform that is capable of executing more complex tasks and running
ROS\@. More specifically, our requirements are as follows:

\begin{itemize}
    \item the price should be in the order of 200  \euro{};
    \item the computational power should be high enough to perform localisation and basic computer vision;
    \item the operating system should be Linux;
    \item hardware extensions with additional sensors and communication
        interfaces should be possible;
    \item the assembly time should not exceed two person-hours;
    \item the platform should also be suited for other research activities in
        our group.
\end{itemize}

We identified multiple platforms that fulfil these requirements and eventually
selected the Balboa robots from Pololu~\cite{noauthor_pololu_nodate}. One of
their advantages for our purpose is a good state estimation capability due to an
integrated inertial measurement unit and quadrature encoders connected to
the motors. The computing power is provided by a RaspberryPi 3B+. Their
self-balancing behaviour is visually attractive, especially when multiple robots
operate next to each other. 

The official software library for the low-level controller (LLC) provided by the
vendor has been tuned to our needs. We changed the communication interface
between the computing board and the LLC to a serial connection using the UART
(universal asynchronous receiver-transmitter) protocol. This enables us to
remotely reprogram the microcontroller and implement a protocol for robot
control and readout of measurements. We have tuned the controller responsible
for balancing to make it work with additional load.  All modifications are
published in the GitLab repository~\cite{url_balboa_llc}. Finally, to visualise
the phase of a swarmalatorbot, we attached a strip of Neopixel LEDs to each
robot.

\subsection{Swarmalatorbots in ROS\,2}

We implemented the swarmalatorbots in ROS\,2 (Bouncy Bolson release)
using eProsima Fast RTPS (Real Time Publish Subscribe) as communication
middleware. ROS\,2 has been chosen despite its early development stage due to its
distributed nature and configurable communication middleware. Multihop
communication is provided by a Babel routing~protocol.

The original swarmalator model assumes that the interactions between oscillators
take place continuously. Robots however can exchange only a limited number of
messages at discrete points in time. Our swarmalatorbots periodically publish
their states (i.e.\ messages containing their identifier, phase and position) to
all other robots with a configurable frequency. Each swarmalatorbot gathers the
received messages and stores the last received update for every robot. The new
control variables are calculated from the model (Box~\ref{box}) by each
swarmalatorbot when it receives information about its position. Between two
published states, control variables might be updated multiple times. The
implementation of the swarmalatorbot can run both in the simulation environment and on the robots in the experiment. For the simulation of more
than $20$ entities the message passing is substituted with intra-process
communication to speed up computations. In order to make the development fast
and the results easy to present, we have implemented a module for
\textit{visualisation}. It listens to messages passed between swarmalatorbots
and renders images. It enables us to visualise both the simulation and the real
robots using the same software. 

\subsection{Localisation}

Swarmalatorbots need to know their positions in a global reference frame.
Outdoors they could acquire their position with GPS, but as our proof of concept
operates indoors, they use a \textit{motion capture system}
(OptiTrack~\cite{noauthor_optitrack_nodate}). To enable future applications of
the model it should run on different robot platforms and in various
environments, both indoor and outdoor. To assure that the robots use the motion
capture system in the same way as they would use GPS, each robot acquires only
its own position from the system, updates its own state and transmits the
position and phase to other~robots.  For the simulation, an external node
calculates positions based on a physical movement model. This approach enables
us to emulate the positioning system.

\subsection{Wireless Communication}

A swarmalatorbot has to exchange its state information with all other
swarmalatorbots and to receive information about its position from OptiTrack.
 Wi-Fi is used for both tasks due to its ease of configuration,
availability and high throughput. Although the wireless modules on RaspberryPi
support 802.11ac, we use by default 802.11bg for the sake of
compatibility with other devices.  The ad-hoc mode is used instead of the
infrastructure mode to make the system robust against failure or loss of some
nodes. Although 802.11 is not optimised for small
packets~\cite{jun_theoretical_2003} used by our application, its performance is
sufficient for the experiments. The communication
overhead will be analysed in future~work.

\subsection{Software Deployment for Multi-Robot Systems}

An automated deployment process is important in swarm robotics since setting up
each of the robots individually is a time-consuming task. As there is no
solution that fulfils our requirements, we  developed a custom generic framework
based on Ansible~\cite{hochstein_ansible_2017}. It enables us to remotely
install, start and stop a program on a selected subset of robots at the same
time. The deployment can be made independent of the robot platform and
programming language. It is possible to specify robot groups (e.g.\ depending on
hardware configuration) on which different versions of software are deployed.
The solution was used for the first time for this paper and is published in the
GitLab repository~\cite{url_ansible_playbooks}.

\begin{figure*}[!t]
    \centering
    \begin{subfigure}[]{0.19\linewidth}
        \includegraphics[width=\textwidth]{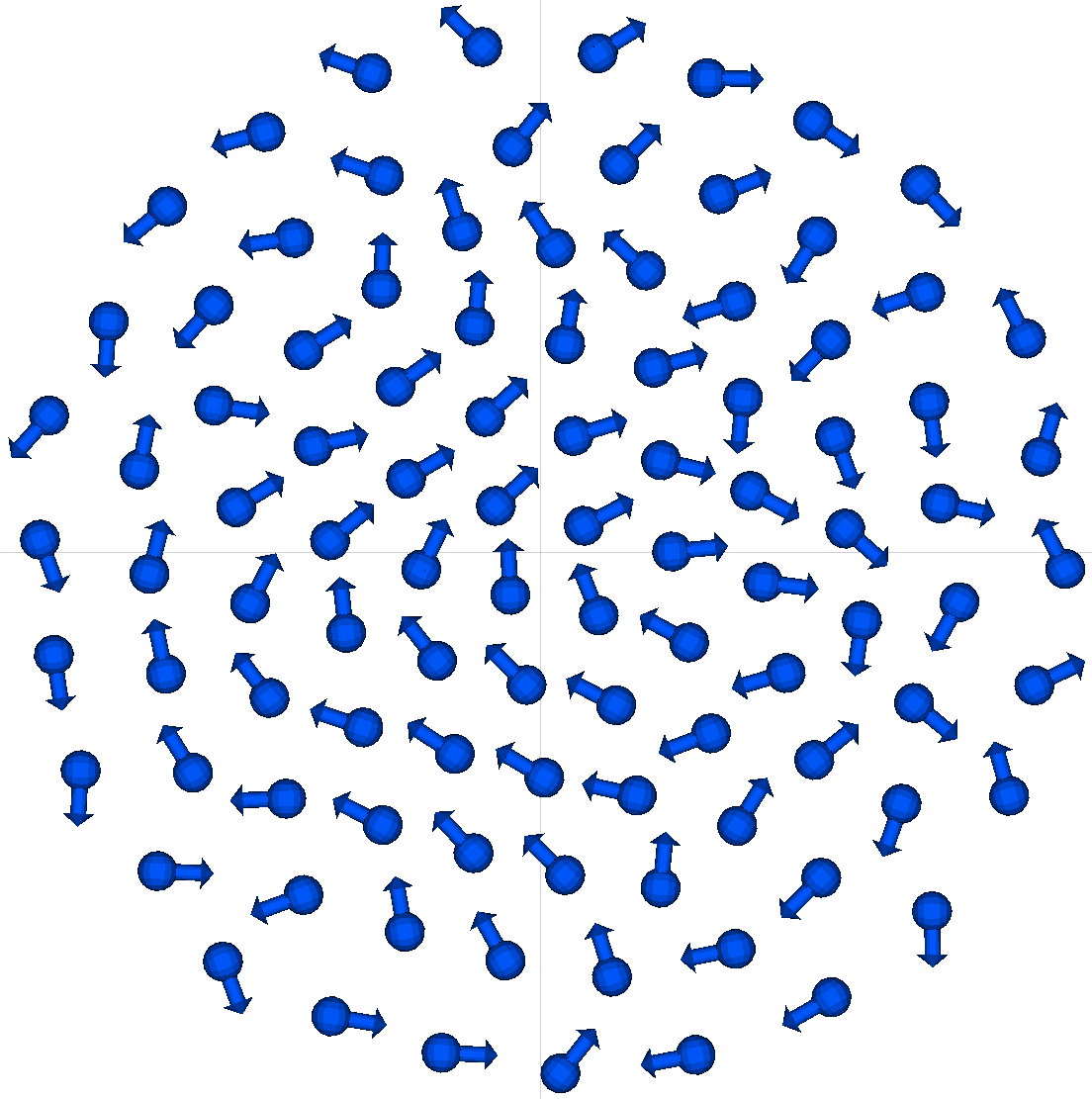}
        \caption{Static sync \\ $(J, K, P) = (0.1, 1, 0)$.}\label{fig:sim-sync}
    \end{subfigure}
    \begin{subfigure}[]{0.19\linewidth}
        \includegraphics[width=\textwidth]{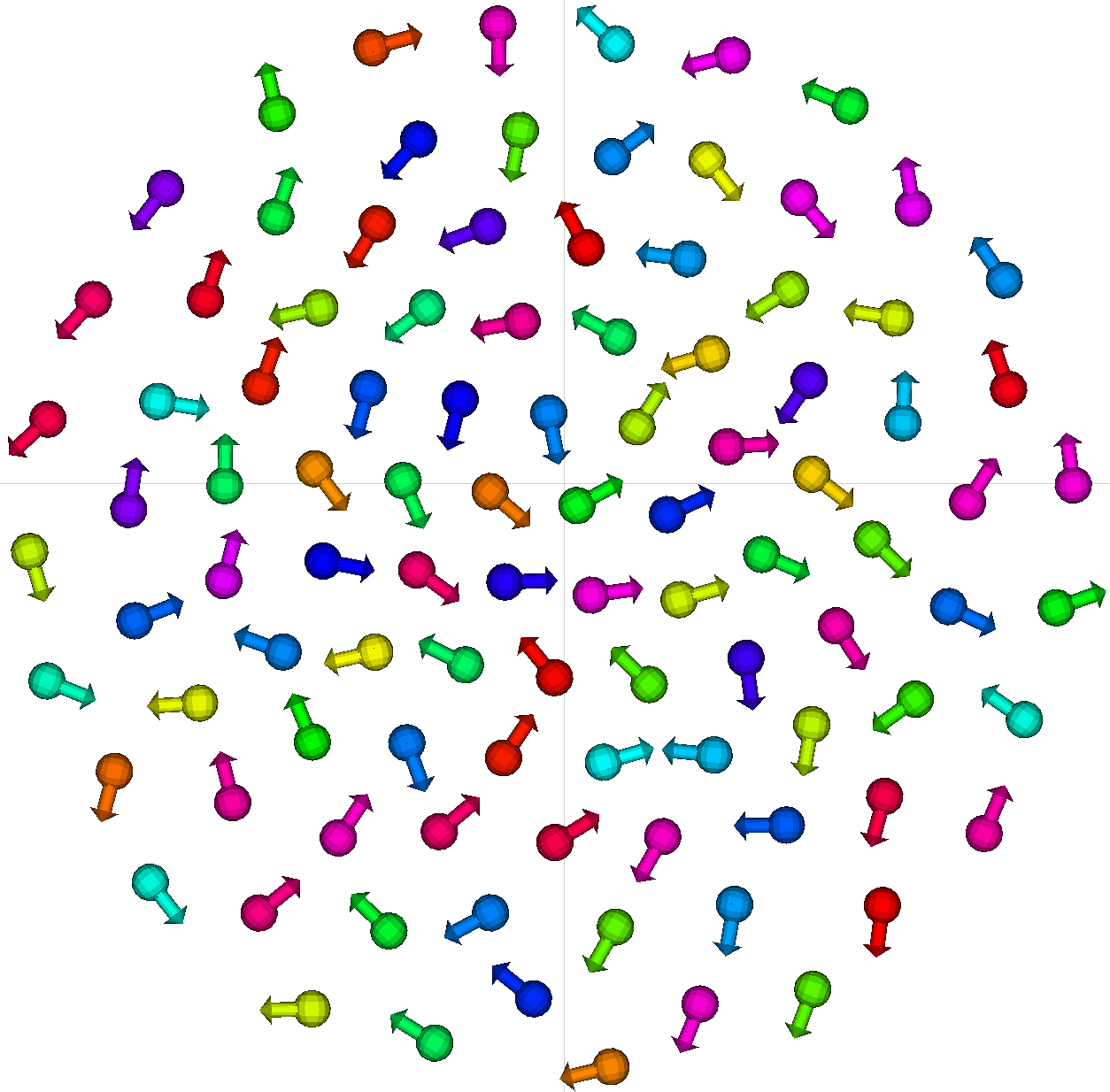}
        \caption{Static async \\ $(J, K, P) = (0.1, -1, 0)$.}\label{fig:sim-async}
    \end{subfigure}
    \begin{subfigure}[]{0.2\linewidth}
        \includegraphics[width=\textwidth]{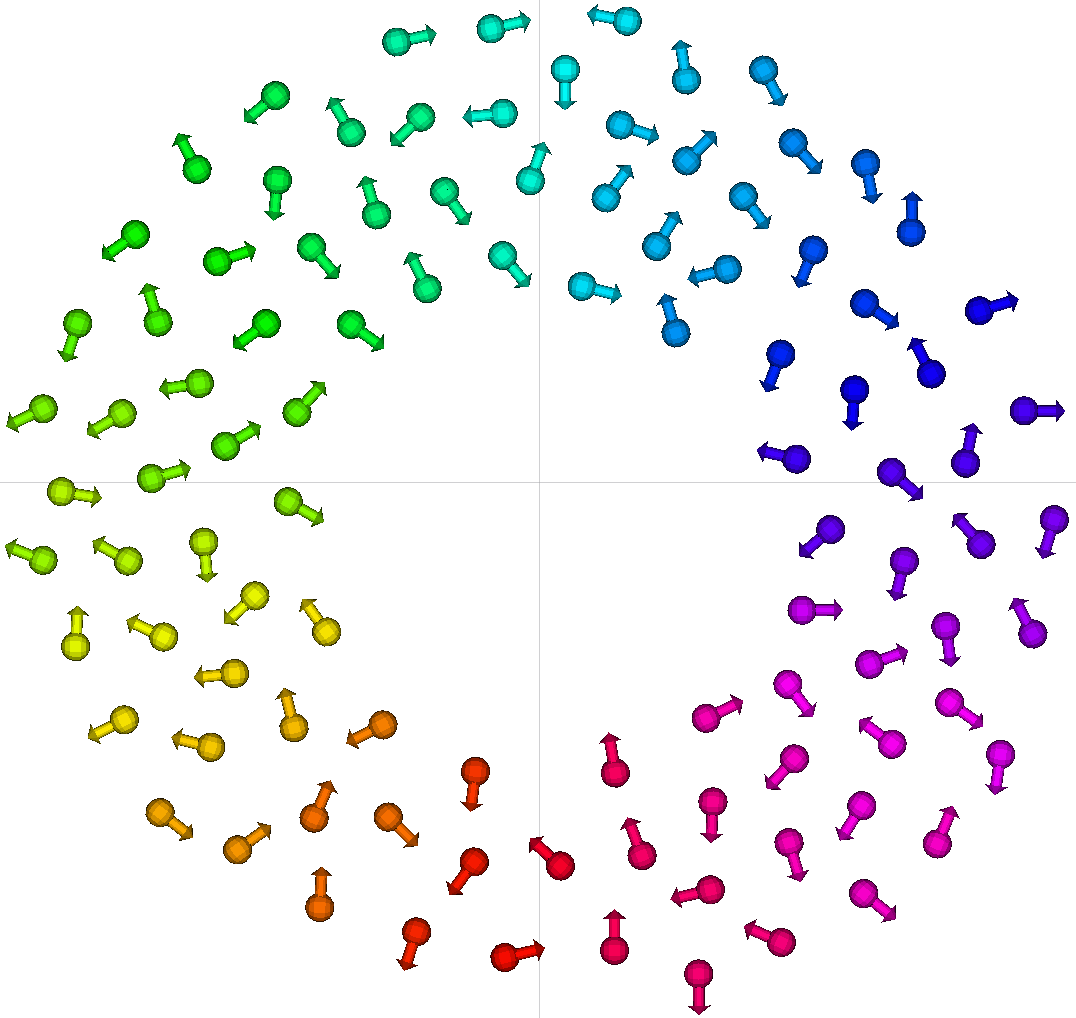}
        \caption{Static phase wave \\ $(J, K, P) = (1, 0, 0)$.}\label{fig:sim-swave}
    \end{subfigure}
    \begin{subfigure}[]{0.19\linewidth}
        \includegraphics[width=\textwidth]{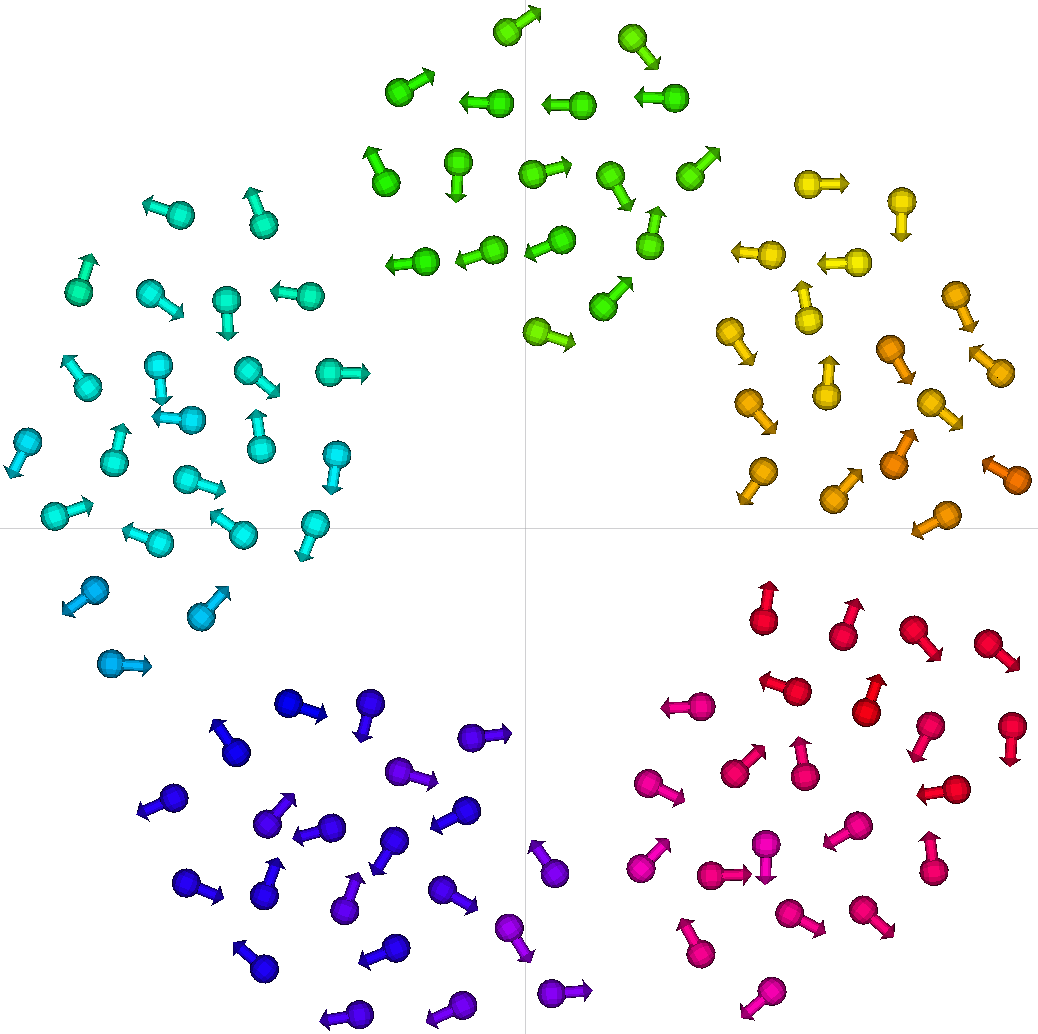}
        \caption{Splintered phase wave \\ $(J, K, P) = (1, -0.1, 0)$.}\label{fig:sim-splinterred}
    \end{subfigure}
    \begin{subfigure}[]{0.205\linewidth}
        \includegraphics[width=\textwidth]{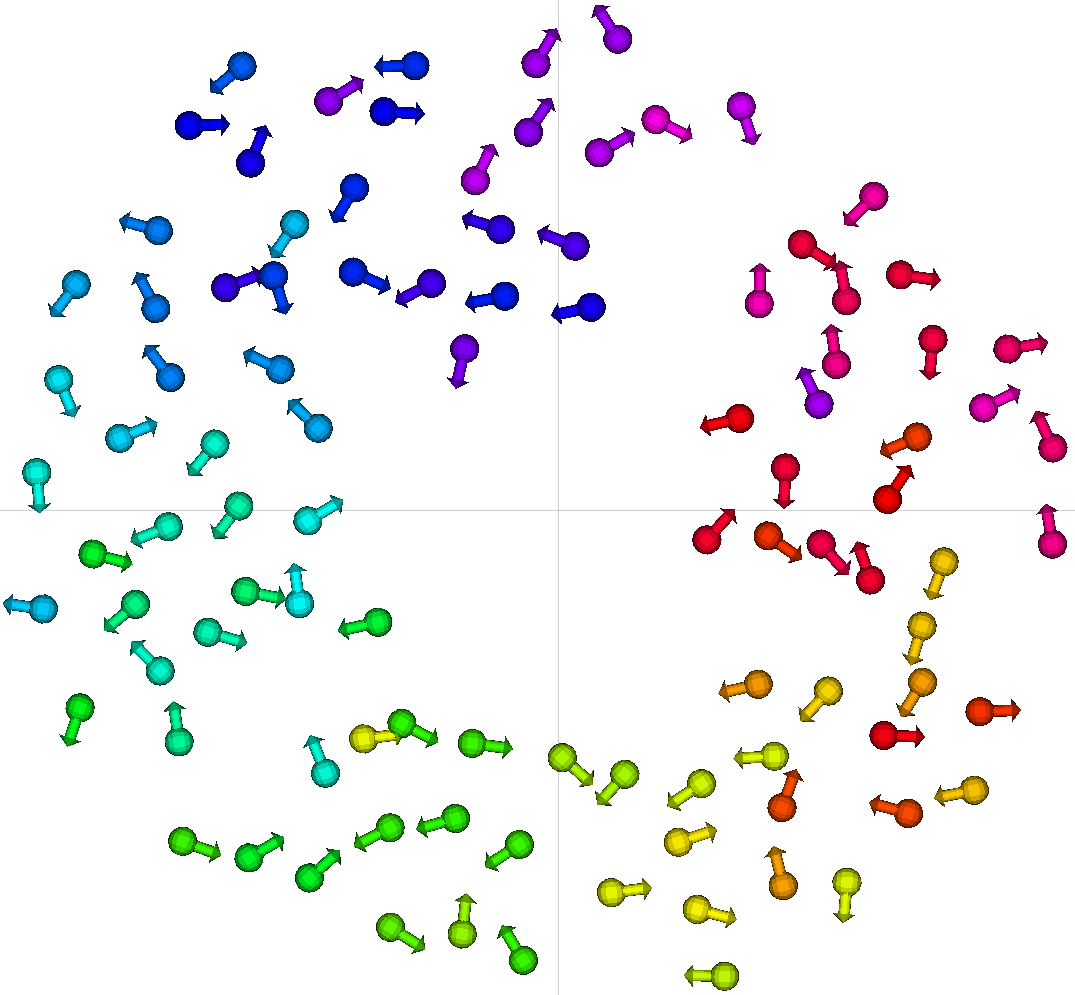}
        \caption{Active phase wave \\ $(J, K, P)=(1, -0.3, 0)$.}\label{fig:sim-awave}
    \end{subfigure}
    \caption{Patterns formed by swarmalators with movement constraints ($N=100$
    entities, time step $dt = 0.05$).}\label{fig:ext-sim}
\end{figure*}

\begin{figure*}[th]
    \centering
    \begin{subfigure}[]{0.19\linewidth}
        \includegraphics[width=\textwidth]{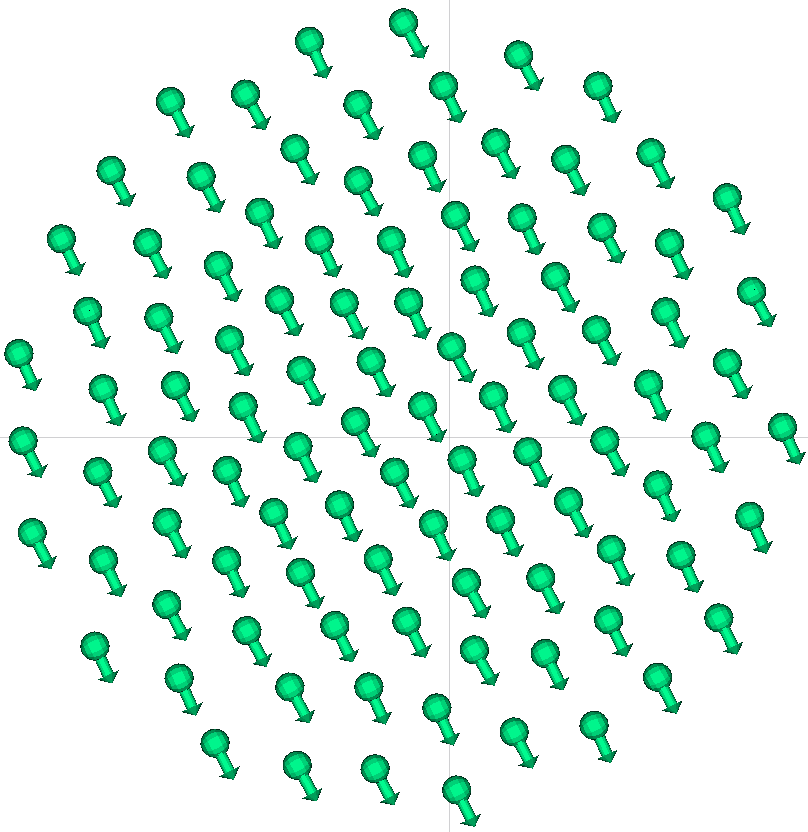}
        \caption{Static sync \\ $(J, K, P)=(0.1, 1, 0.1)$.}\label{fig:sim-sync-a}
    \end{subfigure}
    \begin{subfigure}[]{0.2\linewidth}
        \includegraphics[width=\textwidth]{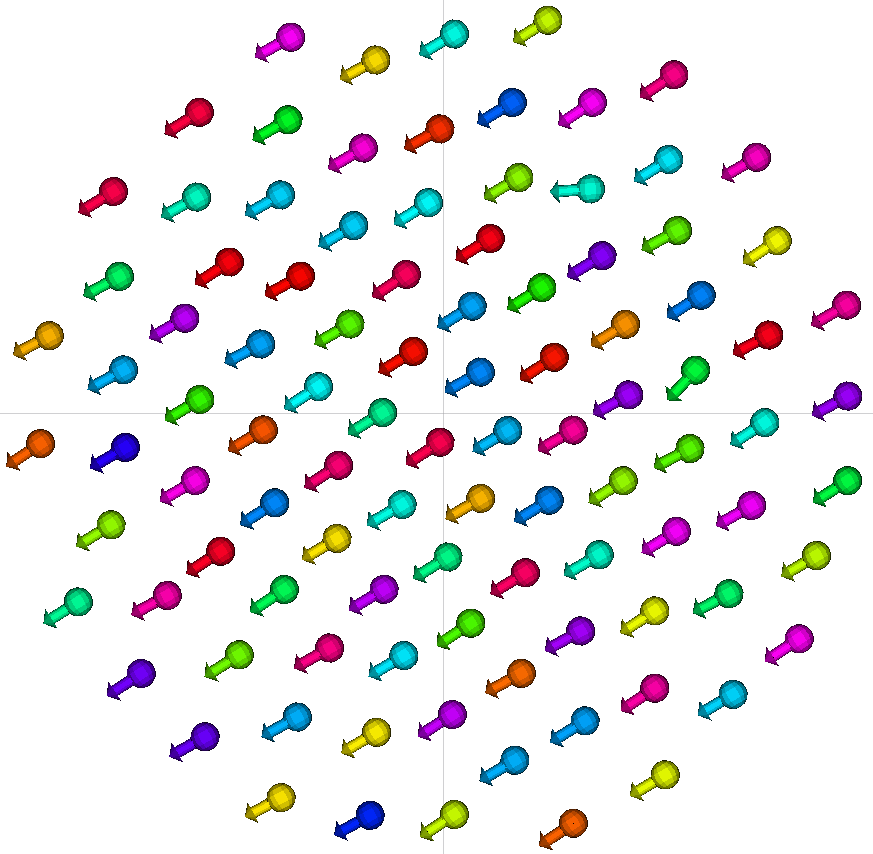}
        \caption{Static async \\ $(J, K, P)=(0.1, -1, 0.1)$.}\label{fig:sim-async-a}
    \end{subfigure}
    \begin{subfigure}[]{0.22\linewidth}
        \includegraphics[width=\textwidth]{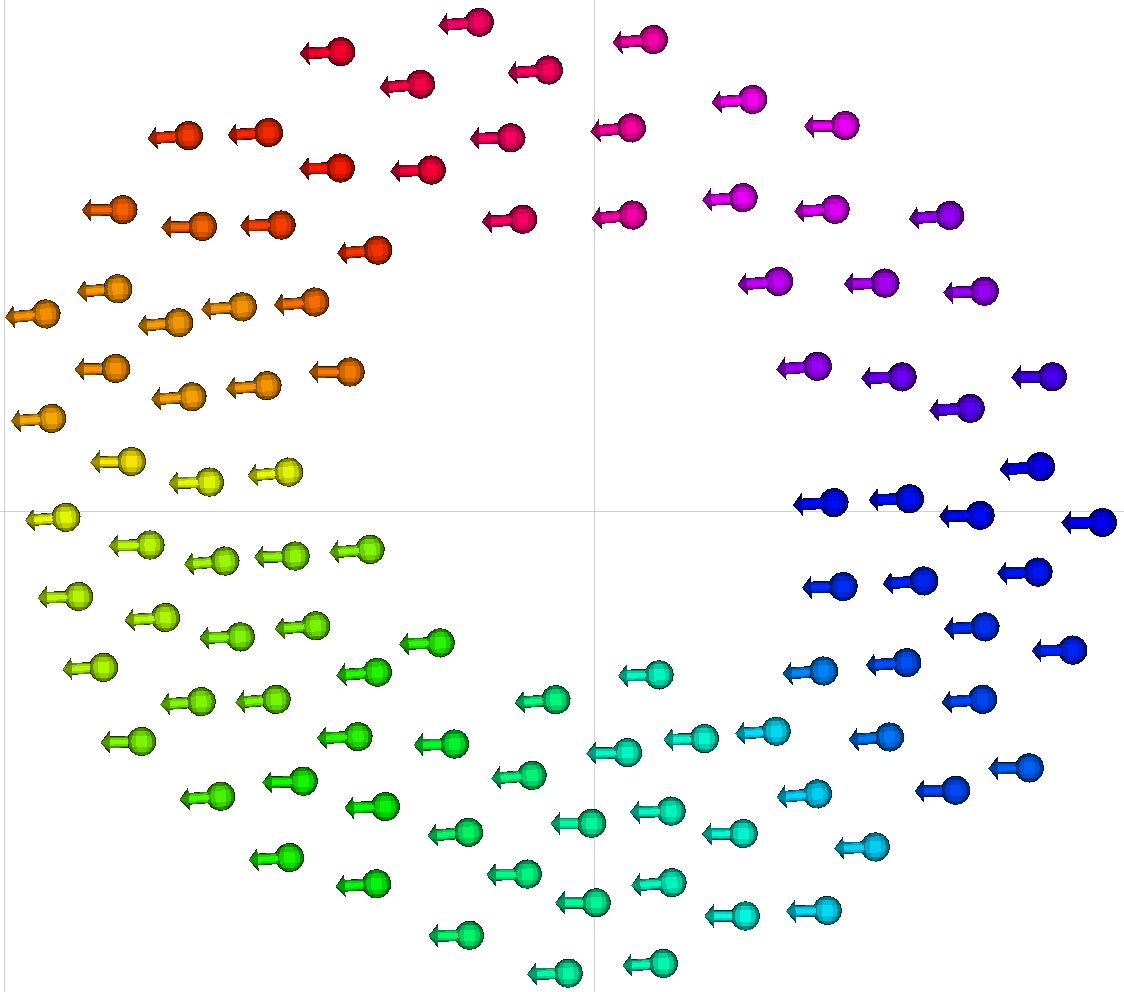}
        \caption{Static phase wave \\ $(J, K, P)=(1, 0, 0.1)$.}\label{fig:sim-swave-a}
    \end{subfigure}
    \caption{Static patterns formed by swarmalators with movement constraints
    and orientation alignment ($N=100$, $dt = 0.05$).}\label{fig:align-sim}
\end{figure*}

\section{Resulting Space-Time Patterns}\label{sec:results}

We now study the behaviour of $N$ swarmalatorbots via ROS\,2 simulations and
experiments with our robotic platform. After initial placement, each entity acts
as a swarmalator and interacts with all other entities. As time progresses, each
entity changes its state in terms of position $\x$, orientation $\orient$ and
phase $\phase$. After some time, a snapshot of the resulting pattern is taken.
The orientation of an entity is visualised by an arrow and its phase by a
colour. All tests are made with the natural frequency $\omega_i = 0\:\forall i$. 

\subsection{Simulation Results}

Simulations are conducted for $N=100$ entities, whose initial placement is
sampled from a uniform random distribution in a square area with $x, y \in
\interval{-1}{1}$ length units.
As a first step, we simulate the original swarmalator model for mutual
validation. All five patterns introduced in~\cite{okeeffe_oscillators_2017} are
qualitatively reproduced.  Next, we simulate the model suited for robots in
order to check whether the modifications influence the capability to form
patterns. \Cref{fig:ext-sim} shows some examples of patterns achieved without
alignment of orientations ($P\!=\!0$). We conclude that it is possible to obtain
all the patterns in this robot model. 

The parameter set $(J, K)=(0.1, 1)$ yields a pattern with entities
synchronised and regularly distributed inside a circle.  The set $(J, K)=(0.1,
-1)$ gives a similar pattern but keeps the nodes in asynchrony. For $(J, K)=(1,
0)$, all entities are regularly distributed on a ring, where the ones with
similar phases are close to each other. The last two states are non-stationary.
For $(J, K)=(1, -0.1)$, the entities create disconnected clusters of similar
phases around a ring and keep moving inside these clusters; this is called the splintered phase wave~\cite{okeeffe_oscillators_2017}. For $(J, K)=(1,
-0.3)$, the entities move around the ring while their phases oscillate; this is
called the active phase wave~\cite{okeeffe_oscillators_2017}.

We observed that\,---\,even with static patterns\,---\,the swarmalatorbots are not
fully stationary but that their positions slightly oscillate. This behaviour can
occur due to the temporal discretisation of the model and because the simulation
is distributed, i.e.\ the communication between entities is asynchronous and may
suffer from message loss. These phenomena can lead to disturbances in the
patterns. 

Some patterns also emerge if we include orientation alignment ($P\!>\!0$).
\Cref{fig:align-sim} shows examples for the three static patterns. The two
non-stationary patterns cannot, in general, be formed with our orientation
alignment.

We observed that the introduction of movement constraints typically slows down
the formation of the patterns. This also means that orientation alignment leads
to even slower convergence. Once the pattern stabilizes, the variance in
position is slightly higher with orientation alignment, as the correction of
disturbances occurs slower.

\subsection{Results with Robotic Prototype}

\newcommand{\expfigwidth}{\linewidth}
\newcommand{\expcolwidth}{0.2\textwidth}

\begin{figure*}[p]
    \centering
    \begin{tabular}[]{m{0.09\textwidth}cccc}
        & \textbf{Robots} & \textbf{Visualisation} & \textbf{Trace: formed
        pattern} & \textbf{Trace: pattern forming} \\

        \textbf{Static sync} &
        \begin{minipage}[c]{\expcolwidth}
            \includegraphics[width=\textwidth,frame]{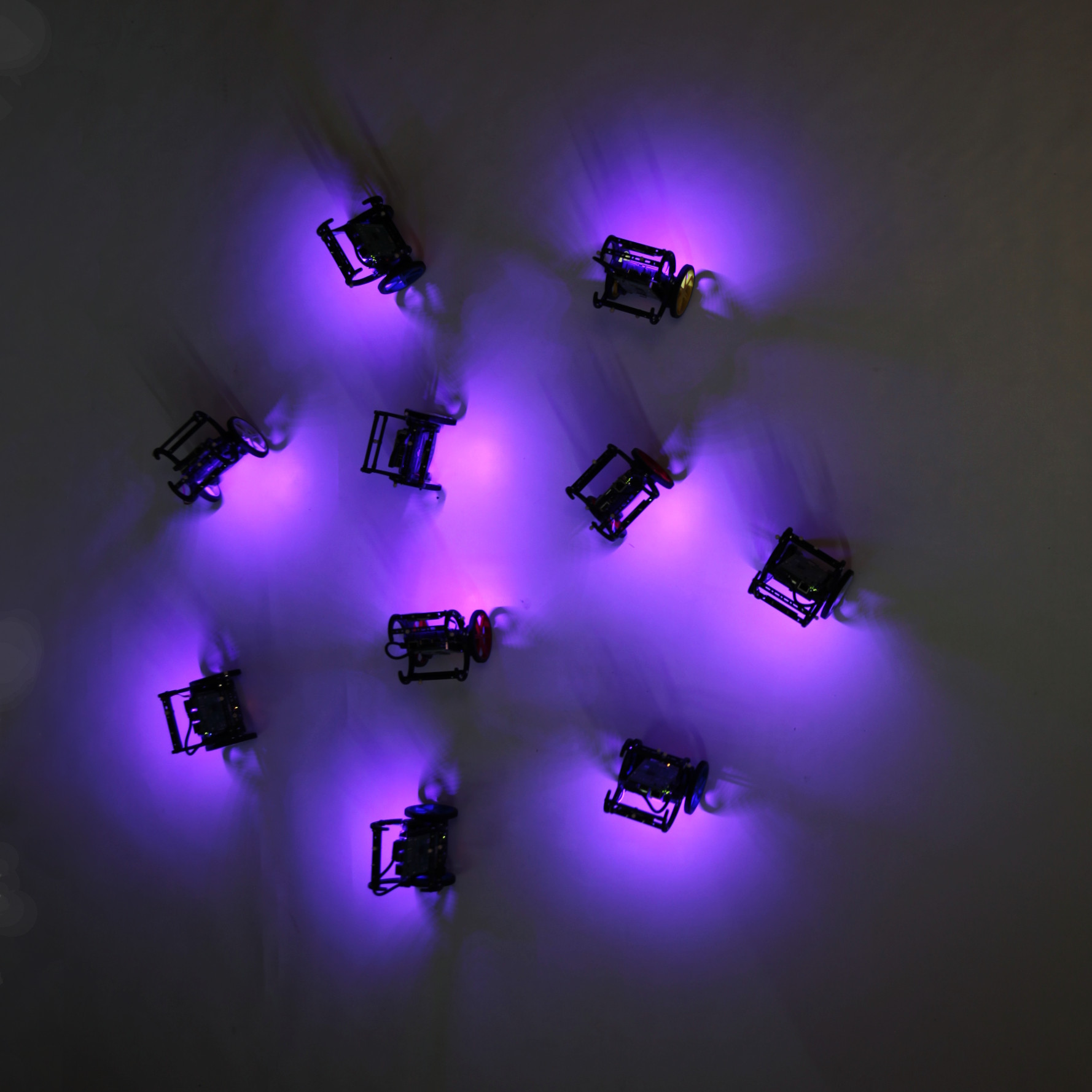}
        \end{minipage} &
        \begin{minipage}[c]{\expcolwidth}
            \includegraphics[width=\textwidth]{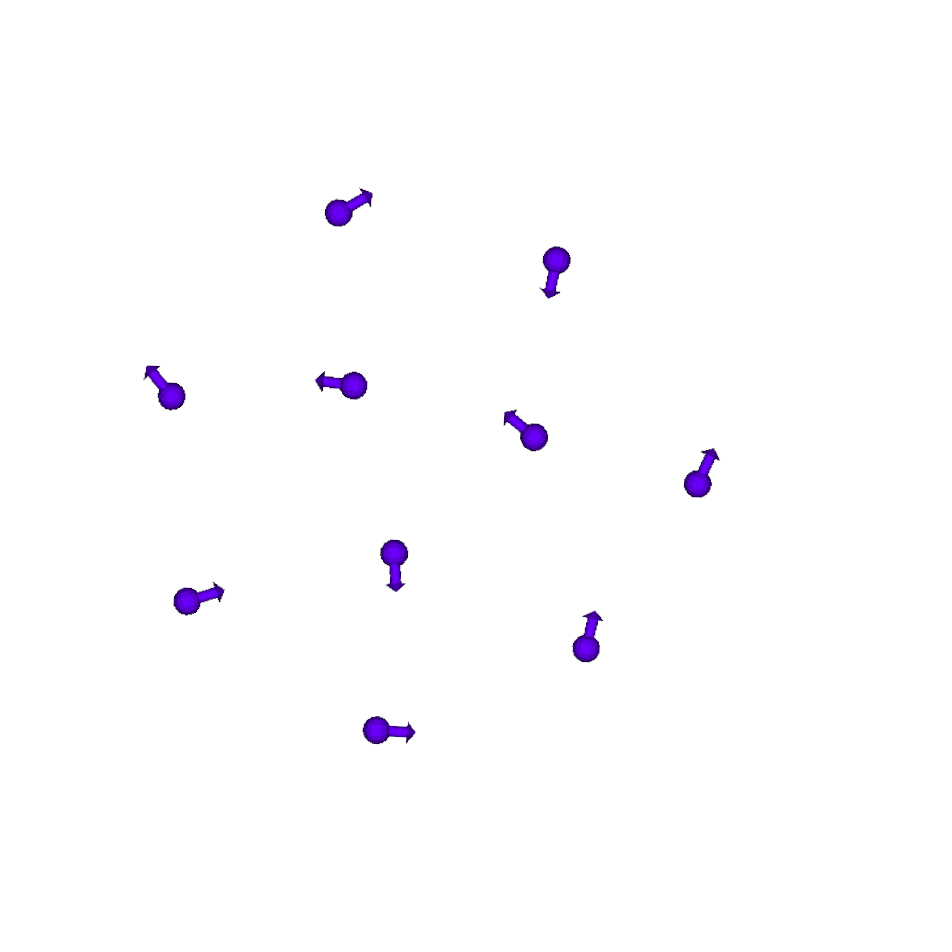}
        \end{minipage} &
        \begin{minipage}[c]{\expcolwidth}
            \includegraphics[width=\textwidth]{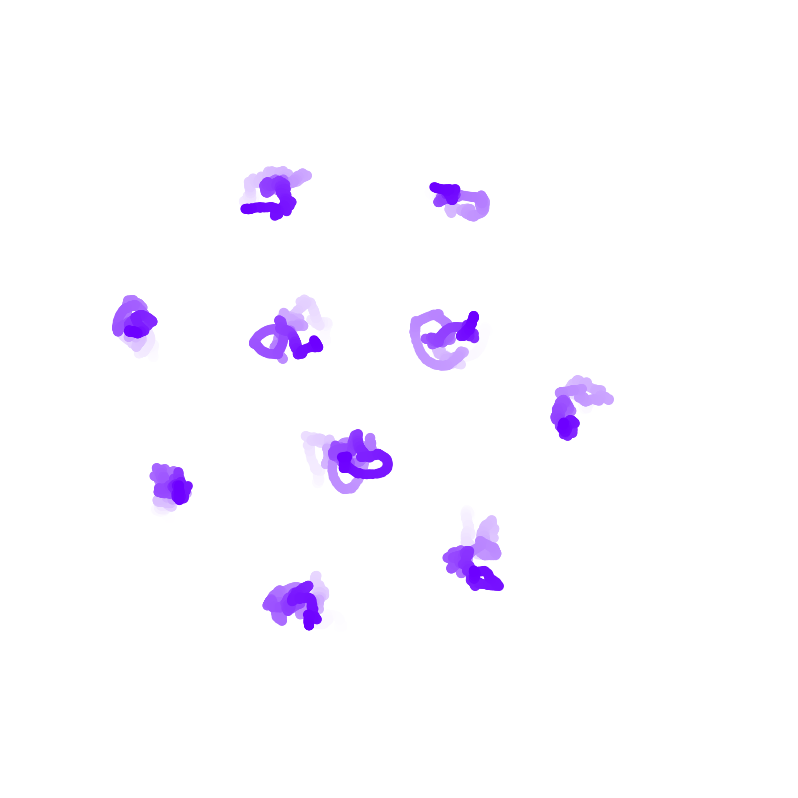}
        \end{minipage} &
        \begin{minipage}[c]{\expcolwidth}
            \includegraphics[width=\textwidth]{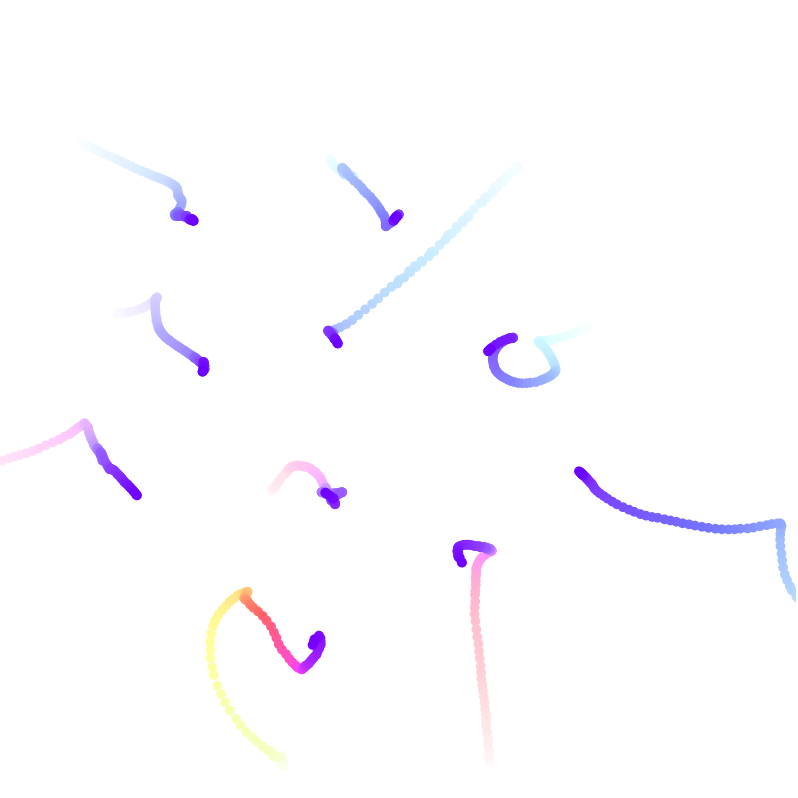}
        \end{minipage} \\

        \textbf{Aligned static sync} &
        \begin{minipage}[c]{\expcolwidth}
            \includegraphics[width=\textwidth,frame]{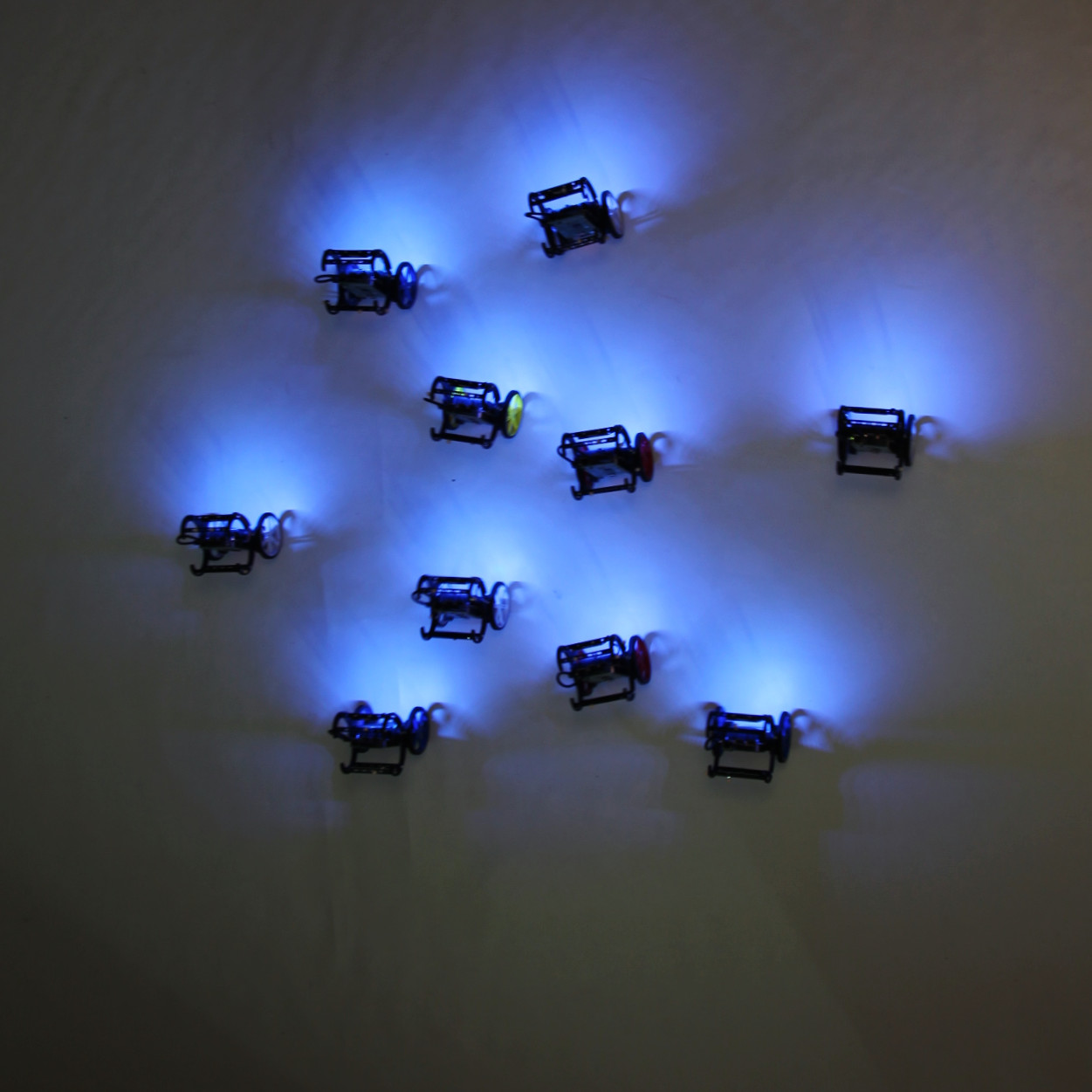}
        \end{minipage} &
        \begin{minipage}[c]{\expcolwidth}
            \includegraphics[width=\textwidth]{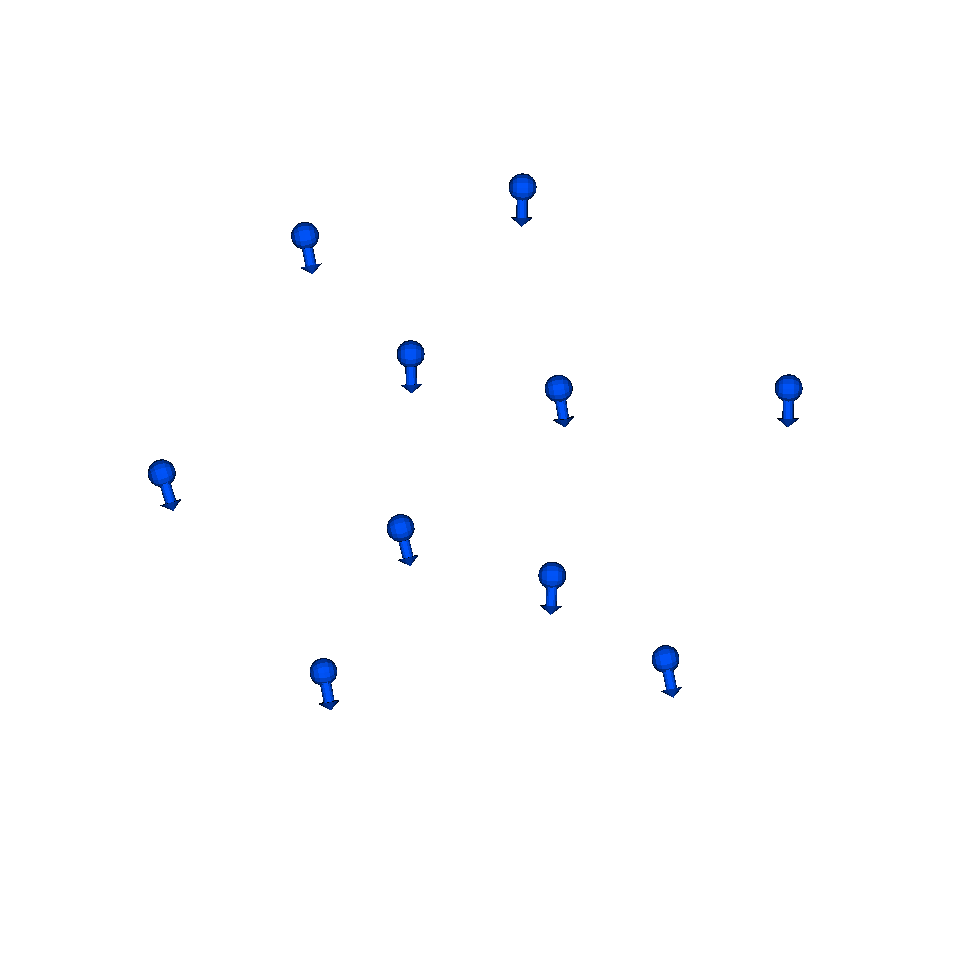}
        \end{minipage} &
        \begin{minipage}[c]{\expcolwidth}
            \includegraphics[width=\textwidth]{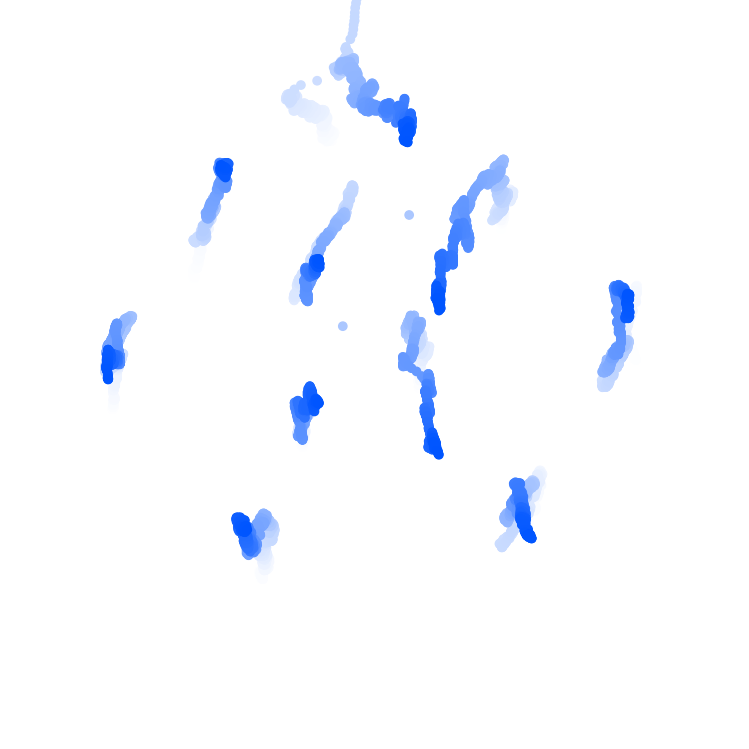}
        \end{minipage} &
        \begin{minipage}[c]{\expcolwidth}
            \includegraphics[width=\textwidth]{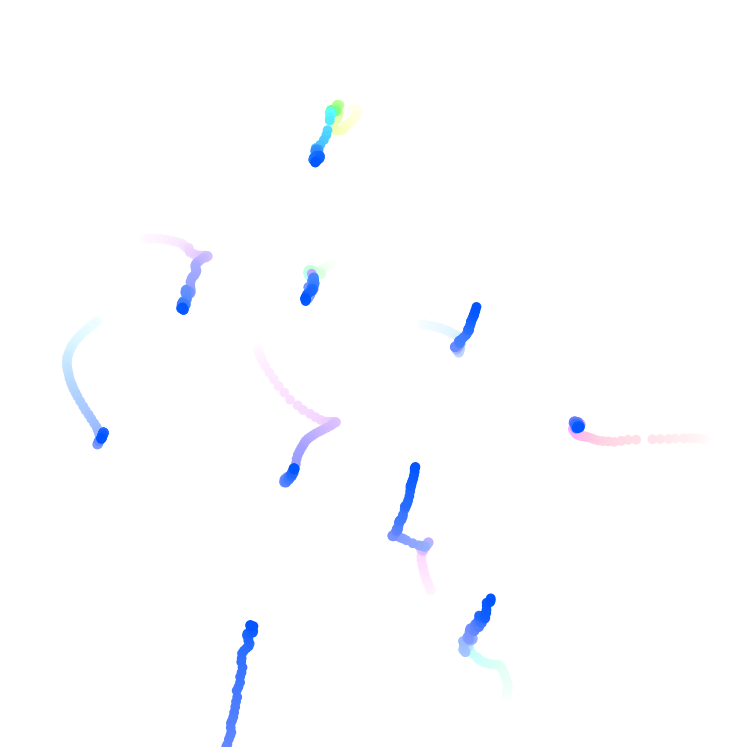}
        \end{minipage} \\
   
        \textbf{Static async} &
        \begin{minipage}[c]{\expcolwidth}
            \includegraphics[width=\textwidth,frame]{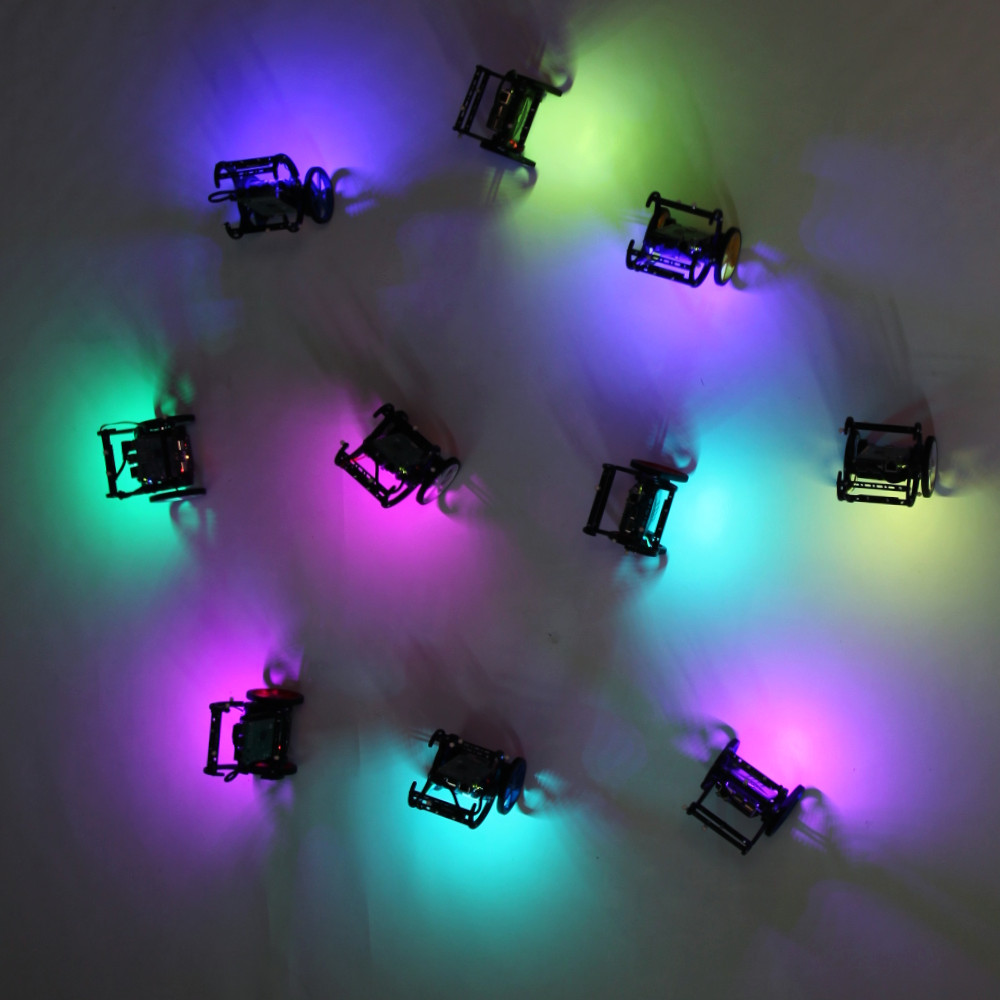}
        \end{minipage} &
        \begin{minipage}[c]{\expcolwidth}
            \includegraphics[width=\textwidth]{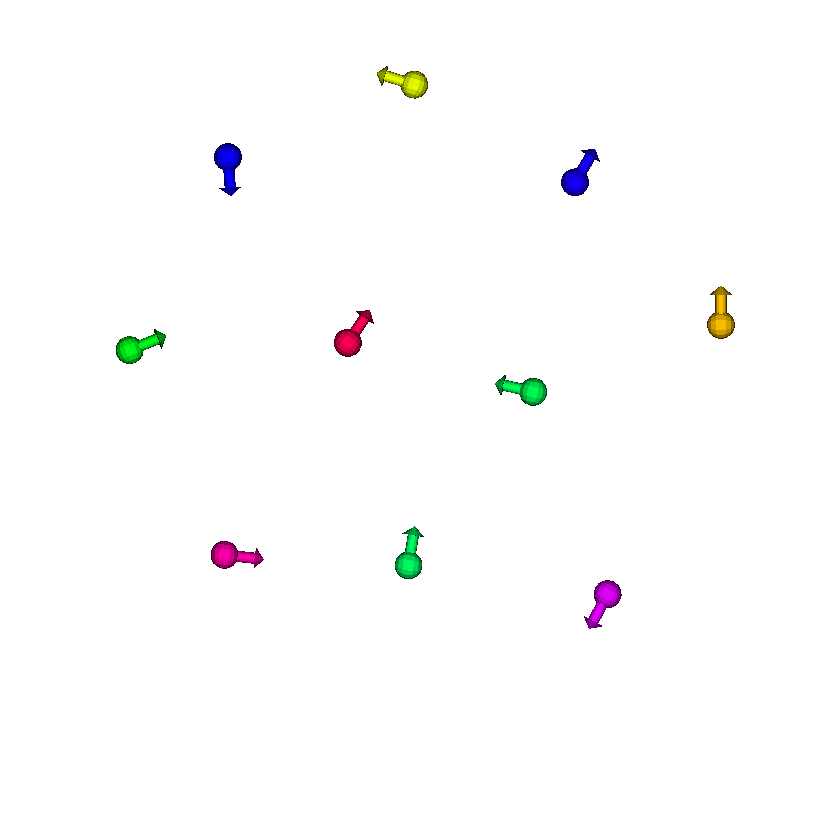}
        \end{minipage} &
        \begin{minipage}[c]{\expcolwidth}
            \includegraphics[width=\textwidth]{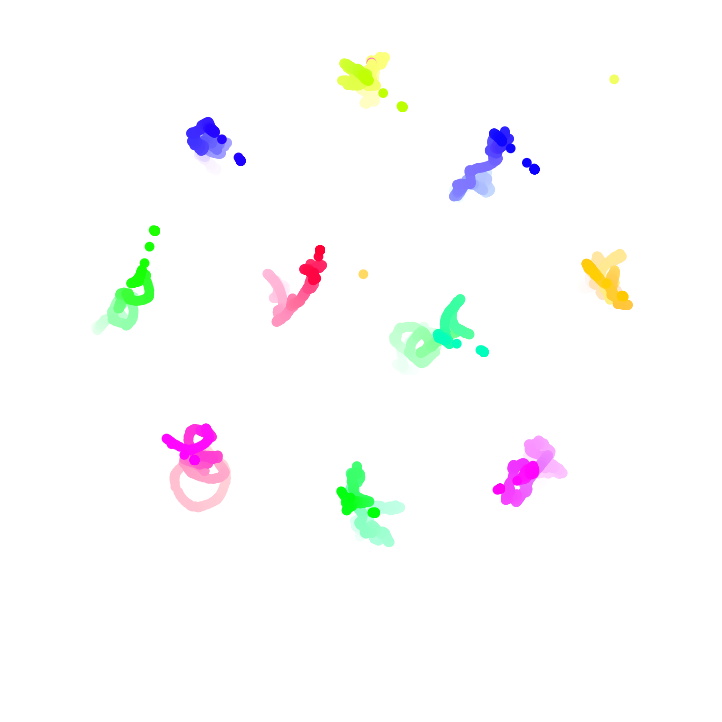}
        \end{minipage} &
        \begin{minipage}[c]{\expcolwidth}
            \includegraphics[width=\textwidth]{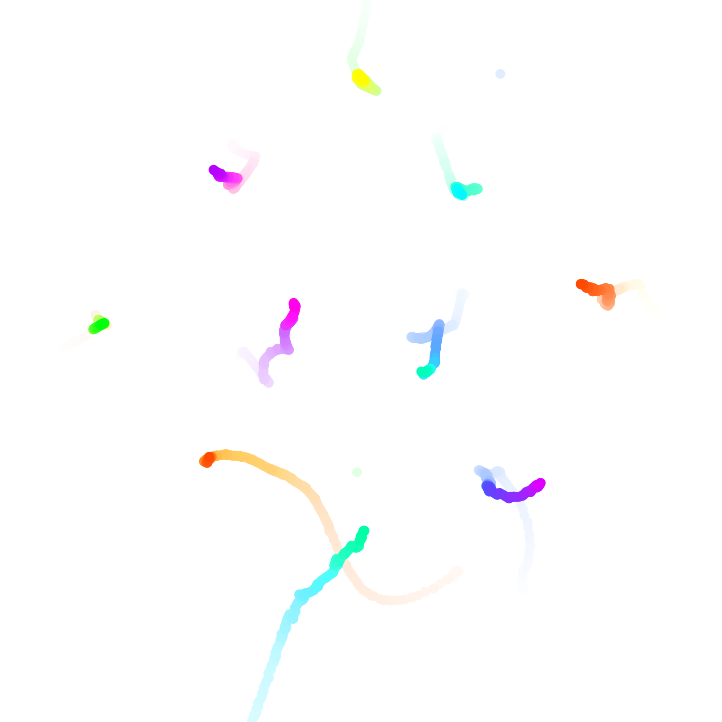}
        \end{minipage} \\
        
        \textbf{Aligned static async} &
        \begin{minipage}[c]{\expcolwidth}
            \includegraphics[width=\textwidth,frame]{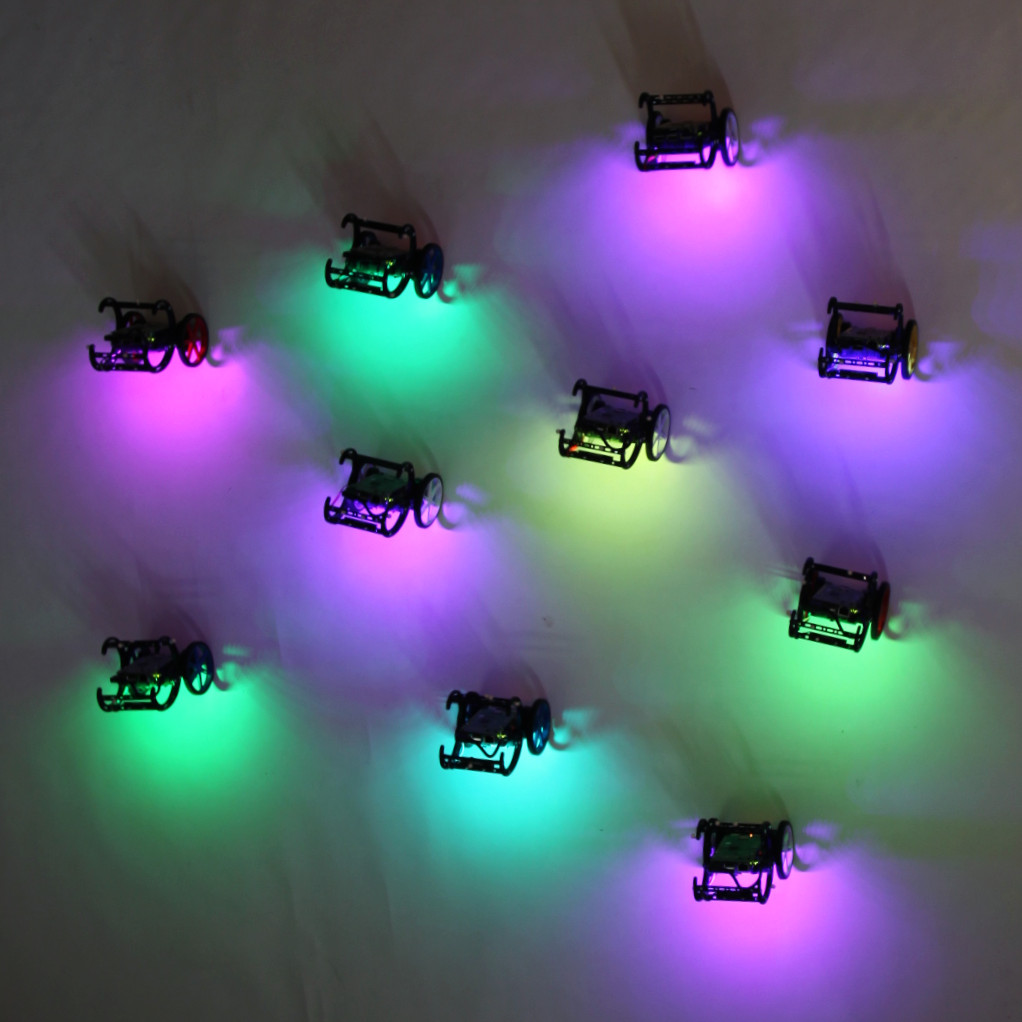}
        \end{minipage} &
        \begin{minipage}[c]{\expcolwidth}
            \includegraphics[width=\textwidth]{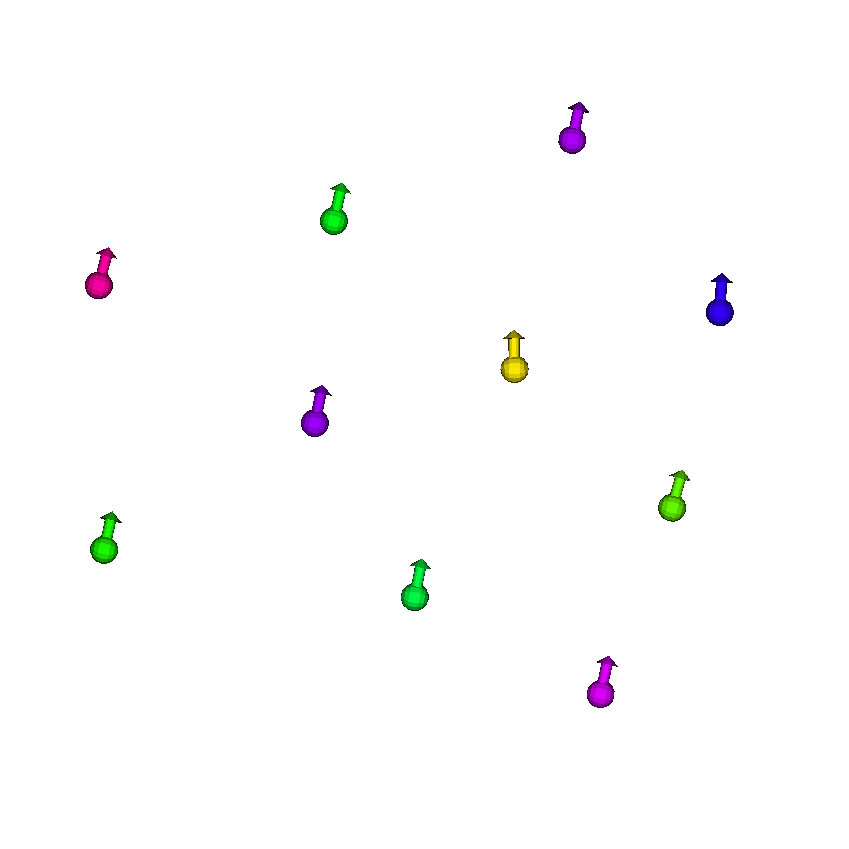}
        \end{minipage} &
        \begin{minipage}[c]{\expcolwidth}
            \includegraphics[width=\textwidth]{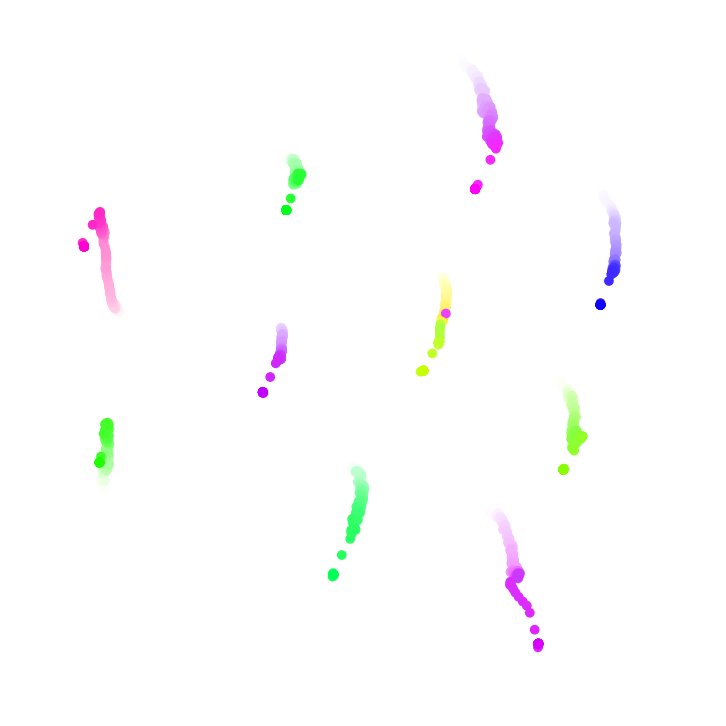}
        \end{minipage} &
        \begin{minipage}[c]{\expcolwidth}
            \includegraphics[width=\textwidth]{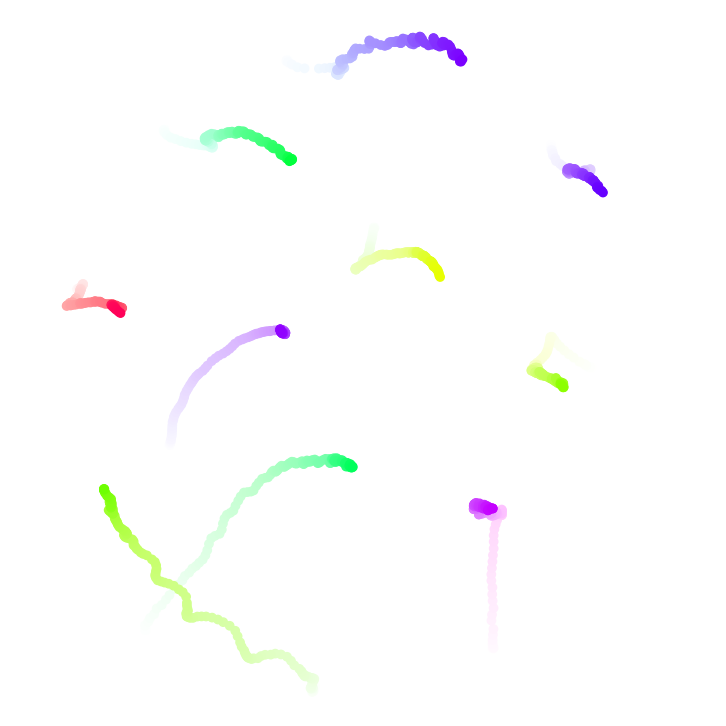}
        \end{minipage} \\
        
        \textbf{Static phase wave} &
        \begin{minipage}[c]{\expcolwidth}
            \includegraphics[width=\textwidth,frame]{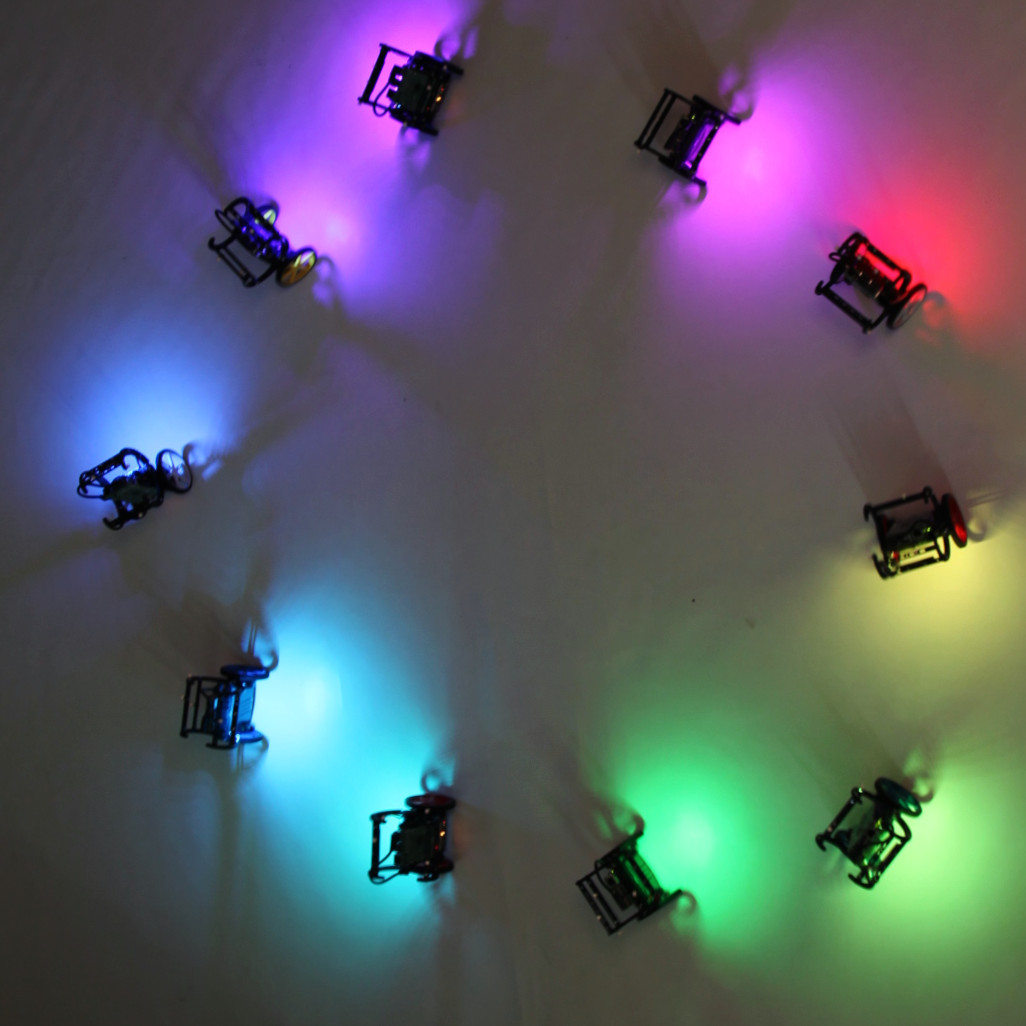}
        \end{minipage} &
        \begin{minipage}[c]{\expcolwidth}
            \includegraphics[width=\textwidth]{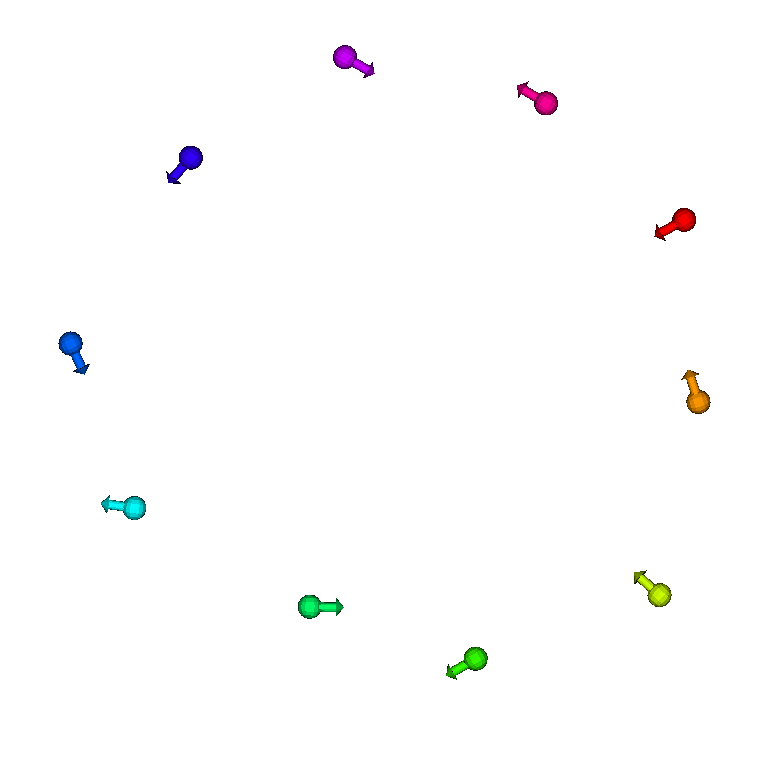}
        \end{minipage} &
        \begin{minipage}[c]{\expcolwidth}
            \includegraphics[width=\textwidth]{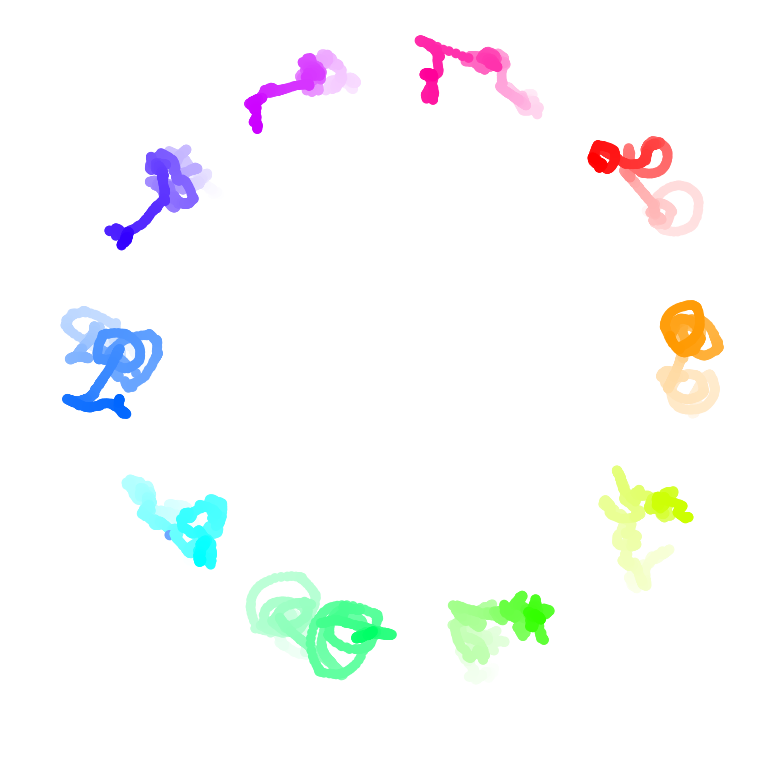}
        \end{minipage} &
        \begin{minipage}[c]{\expcolwidth}
            \includegraphics[width=\textwidth]{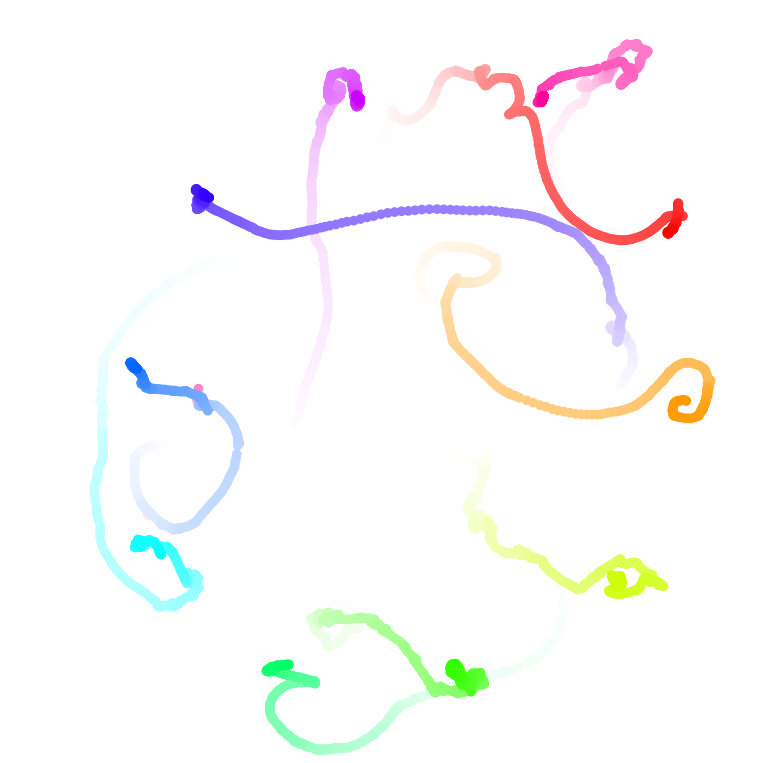}
        \end{minipage} \\

        \textbf{Aligned static phase wave} &
        \begin{minipage}[c]{\expcolwidth}
            \includegraphics[width=\textwidth,frame]{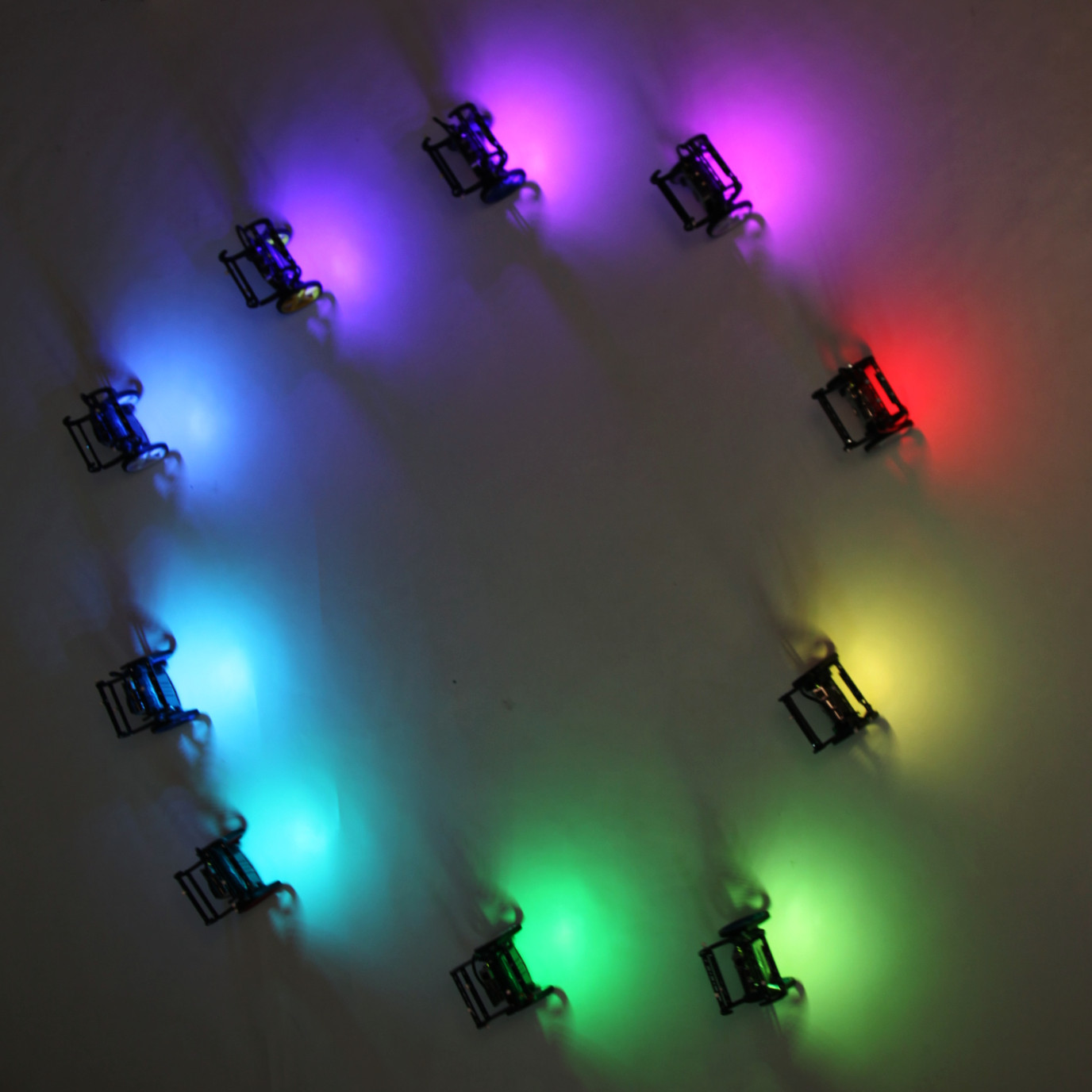}
        \end{minipage} &
        \begin{minipage}[c]{\expcolwidth}
            \includegraphics[width=\textwidth]{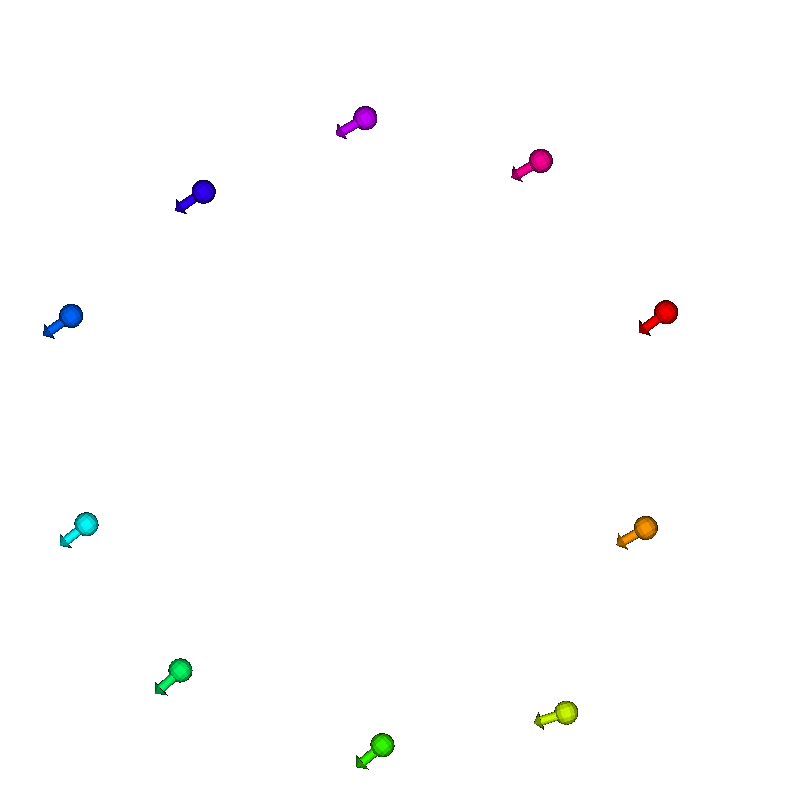}
        \end{minipage} &
        \begin{minipage}[c]{\expcolwidth}
            \includegraphics[width=\textwidth]{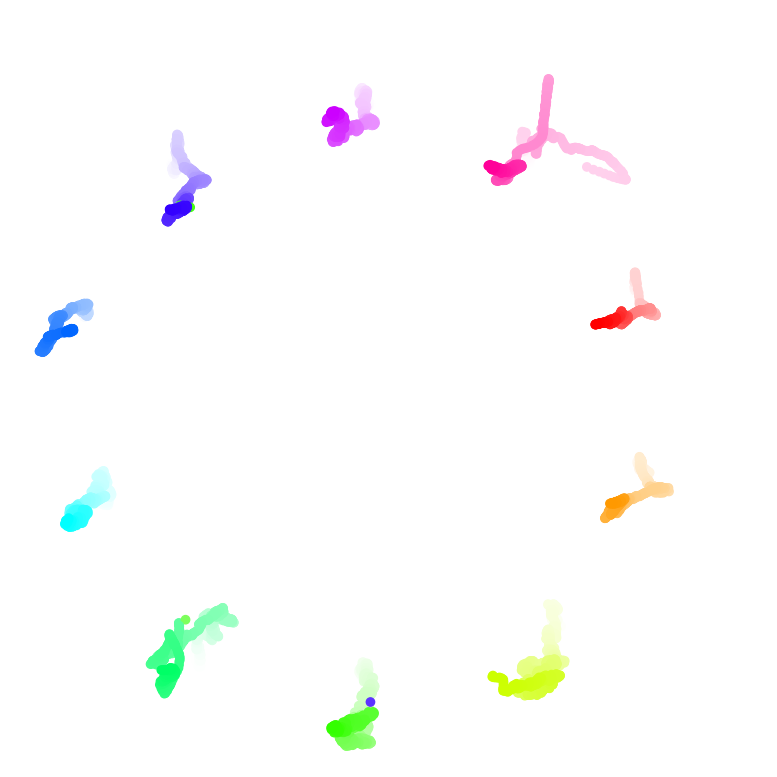}
        \end{minipage} &
        \begin{minipage}[c]{\expcolwidth}
            \includegraphics[width=\textwidth]{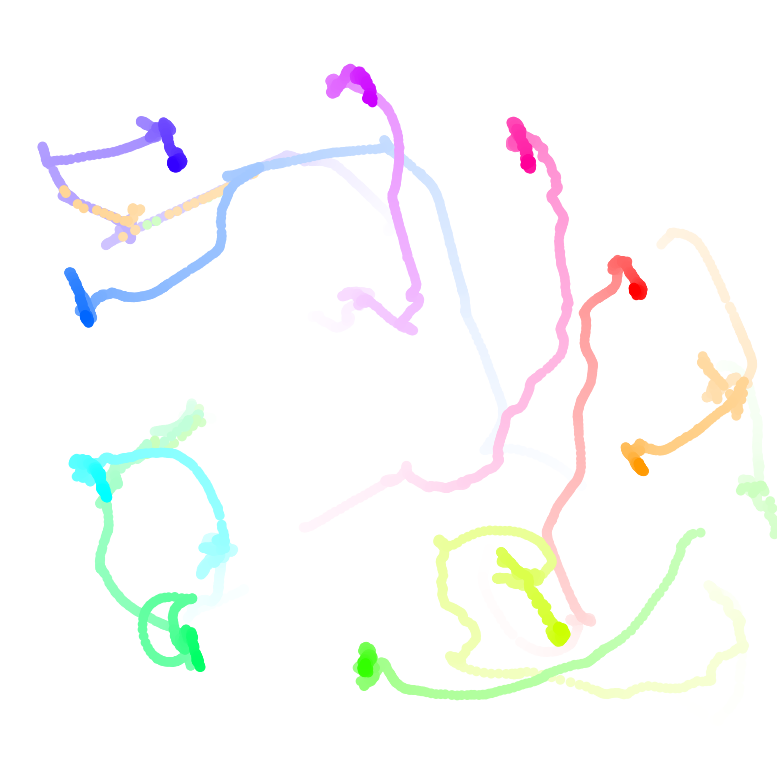}
        \end{minipage} \\
    \end{tabular}
    \caption{Stationary patterns formed by swarmalatorbots.}\label{fig:results-robots}
\end{figure*}

\begin{figure}[]
    \centering
    \begin{subfigure}{0.45\linewidth}
        \includegraphics[width=\textwidth,frame]{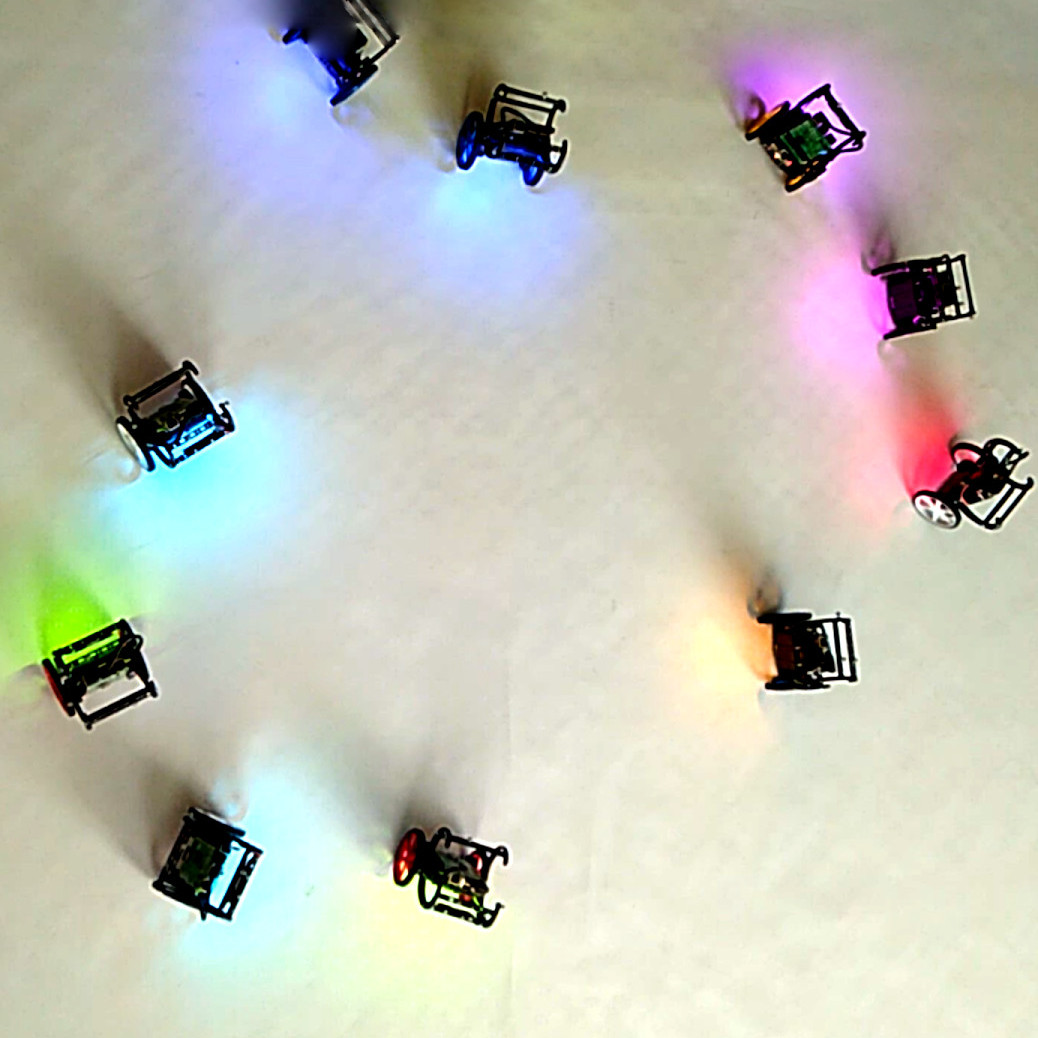}
        \caption{}\label{fig:results-active-photo}
    \end{subfigure}
    \begin{subfigure}{0.45\linewidth}
        \includegraphics[width=\textwidth]{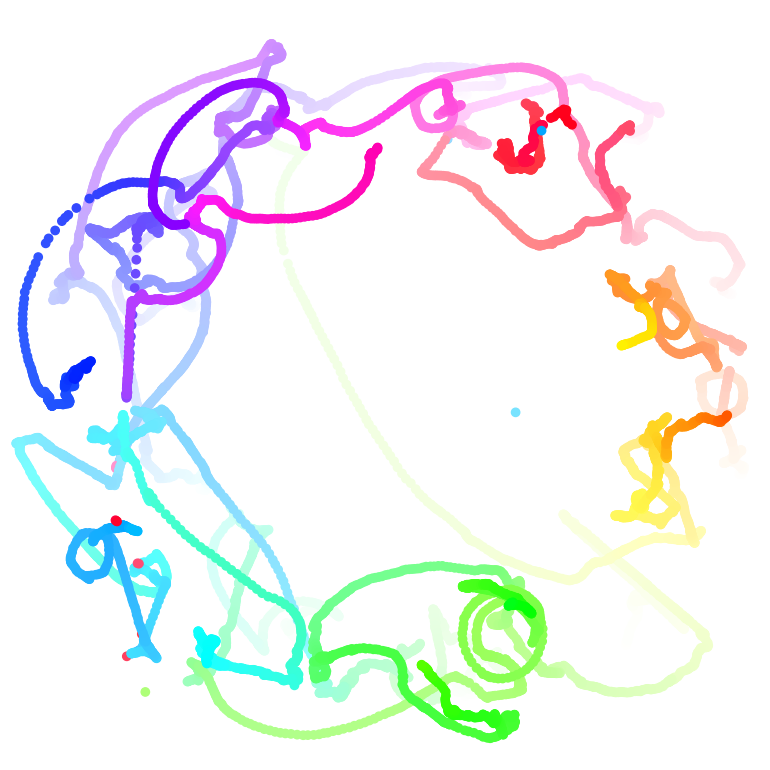}
        \caption{}\label{fig:results-active-trace}
    \end{subfigure}
    \caption{Active phase wave formed by
    swarmalatorbots.}\label{fig:results-robots-active}
\end{figure}

Experiments are made with $N=10$ robots running the model implemented in ROS\,2.
A video about our experiments can be seen on the website~\cite{url_video}.
Example results are presented in
\Cref{fig:results-robots,fig:results-robots-active}. All patterns except the
splintered phase wave are reproduced. The splintered phase wave is not visible
with our setup due to the low number of entities. For all stationary patterns
(\Cref{fig:results-robots}), both versions with and without orientation
alignment are created. Despite the low number of entities, the formed patterns
are clearly visible and can be mapped to the different pattern types.
\Cref{fig:results-robots} contains four figures for each stationary state: the
bird's eye view of the robots forming the pattern (\textit{Robots}), snapshot
from the visualisation tool (\textit{Visualisation}), trace with the formed
pattern, showing that the state is stationary (\textit{Trace: formed pattern})
and the trace of pattern forming (\textit{Trace: pattern forming}).  Older
samples are visualised with a lighter colour.  The active phase wave formed by
robots is presented in~\cref{fig:results-robots-active}; the trace shows the
robots' movement after the pattern has formed.

Disturbances similar to the ones in the simulations can be observed with the real
robots.  They become even more visible in the experiments because of the
variable communication characteristics as well as the robot's inertia and
imperfect state estimation.

\section{Related Work}\label{sec:rel-work}

There is a broad spectrum of scientific work on spatial and temporal coordination techniques that have been applied to multi-robot systems. 

Publications on spatial coordination propose and evaluate methods for pattern formation and control  \cite{suzuki_distributed_1999,hsu_multiagent-based_2005,barnes_unmanned_2007}, swarm navigation with obstacle avoidance~\cite{junior_efficient_2016}, decentralised flocking \cite{turgut_self-organized_2008,ferrante_self-organized_2012}, and dependency of flocking on communication~\cite{hauert11:iros}, to give some examples. All these approaches are without any interaction regarding temporal coordination.

Temporal coordination is often used in engineered systems but seldom coupled with spatial coordination. Algorithms in this domain suited for multi-robot systems are often inspired by phenomena of self-organising synchronisation in nature. Here, entities exchange simple pulses over radio~\cite{brandner_firefly_2016,hong_scalable_2005}, sound~\cite{trianni_self-organizing_2009}, or light~\cite{perez-diaz_firefly-inspired_2015}.
There are only a few approaches combining synchronisation and swarming. They use
synchronisation for swarm coordination but do not couple them in a bidirectional
way. Christensen et al.~propose a method in which robots detect faulty agents
because they have stopped blinking~\cite{christensen_fireflies_2009}. Hartbauer et
al.~describe how light emitted by robots can be used as a guiding
signal~\cite{hartbauer_novel_2007}.

\section{Conclusions and Outlook}\label{sec:concl}

The swarmalator concept for coupled synchronisation and swarming has been
realised and investigated for the first time in a real-world system. We have
adapted and extended the original mathematical model for its use in mobile
robotics and implemented it in ROS\,2. Simulations and experiments with our
Balboa-based platform serve as a proof of concept. The real-world demonstrations reveal some artefacts, e.g.~the  swarmalatorbots slightly oscillate around a given position, even with static patterns. 
Furthermore, we observed that the communication delay has a huge impact on
the patterns stationarity if $\omega > 0$. For example, the static phase wave
starts to rotate. This phenomenon requires further investigation. 

Applications of swarmalatorbots are in
fields with coordinated mobility of multiple entities, such as monitoring,
exploration, entertainment and art. To realise these applications, our future work will use small drones creating three-dimensional swarmalator patterns in outdoor~environments. 

\section*{Acknowledgements}

This work was supported by the Karl Popper Kolleg on {Networked Autonomous
Aerial Vehicles} at the University of Klagenfurt and by the Austrian Science
Fund (FWF) under grant P30012-NBL on self-organising synchronisation. 

\bibliographystyle{IEEEtran}
\balance%
\bibliography{bettstetter,sync-and-swarm}
\end{document}